\documentclass[10pt,journal,compsoc]{IEEEtran}
\IEEEoverridecommandlockouts
\ifCLASSOPTIONcompsoc   
  \usepackage[nocompress]{cite}
\else
  \usepackage{cite}
\fi

\usepackage[utf8]{inputenc}
\usepackage{amssymb}
\usepackage{latexsym}
\usepackage{url}
\usepackage{xcolor}
\definecolor{newcolor}{rgb}{.8,.349,.1}

\usepackage{soul}

\usepackage{epsfig}
\usepackage{amssymb}
\usepackage{amsmath}
\usepackage{amsthm}
\usepackage{bm}
\usepackage{amsfonts}
\usepackage{cases}
\usepackage{esint}
\usepackage{graphicx}
\usepackage{subfig}
\usepackage{algorithm} 
\usepackage{algorithmic} 
\usepackage{multirow} 
\usepackage{amsmath}
\usepackage{xcolor}

\usepackage{makecell}
\usepackage[normalem]{ulem}
\usepackage{overpic}

\newcommand{\secref}[1]{Section~\ref{#1}}

\newcommand{\figref}[1]{Fig.~\ref{#1}}
\newcommand{\eqnref}[1]{Equ.~(\ref{#1})}
\newcommand{\tabref}[1]{Table~\ref{#1}}

\newcommand{\sArt}[0]{state-of-the-art}
\newcommand{\gc}[1]{}

\usepackage{array}
\newcommand{\PreserveBackslash}[1]{\let\temp=\\#1\let\\=\temp}
\newcolumntype{C}[1]{>{\PreserveBackslash\centering}p{#1}}
\newcolumntype{R}[1]{>{\PreserveBackslash\raggedleft}p{#1}}
\newcolumntype{L}[1]{>{\PreserveBackslash\raggedright}p{#1}}

\begin{document}
\title{VecQ: Minimal Loss DNN Model Compression\\With Vectorized Weight Quantization}
\author{Cheng~Gong, 
       Yao~Chen,
        Ye~Lu,
        Tao~Li,
        Cong~Hao,
        Deming~Chen,~\IEEEmembership{Fellow,~IEEE}
\IEEEcompsocitemizethanks{
\IEEEcompsocthanksitem Cheng Gong is with the College of Computer Science, Nankai University, Tianjin, China, and the Tianjin Key Laboratory of Network and Data Security Technology. 
E-mail: cheng-gong@mail.nankai.edu.cn.
\IEEEcompsocthanksitem Yao Chen is with Advanced Digital Sciences Center, Singapore. 
~E-mail: yao.chen@adsc-create.edu.sg.
\IEEEcompsocthanksitem Tao Li is with the College of Computer Science, Nankai University, Tianjin, China, and the Tianjin Key Laboratory of Network and Data Security Technology.
E-mail: litao@nankai.edu.cn.
\IEEEcompsocthanksitem Ye Lu is with the College of Cyber Science, Nankai University, Tianjin, China.\protect 
~E-mail: luye@nankai.edu.cn.
\IEEEcompsocthanksitem Cong Hao is with the Electrical and Computer Engineering, the Grainger College of Engineering, University of Illinois at Urbana-Champaign, IL, USA. 
~E-mail: congh@illinois.edu.
\IEEEcompsocthanksitem Deming Chen is with the Electrical and Computer Engineering, the Grainger College of Engineering, University of Illinois at Urbana-Champaign and Advanced Digital Sciences Center, Singapore. \protect
E-mail: dchen@illinois.edu.}
\thanks{Manuscript received Oct 15, 2019; revised March 08, 2020.}
\thanks{(Corresponding~authors:~Ye~Lu, Tao~Li and Deming Chen.)} 
\thanks{Recommended for acceptance by the SI on Machine Learning Architectures Guest Editors.}
\thanks{Digital Object Identifier no. 10.1109/TC.2020.2995593}
}

\markboth{IEEE Transactions on Computers}%
{Gong \MakeLowercase{\textit{et al.}}: VecQ: Minimal Loss DNN Model Compression\\With Vectorized Weight Quantization}

\IEEEtitleabstractindextext{
\begin{abstract}
Quantization has been proven to be an effective method for reducing the computing and/or storage cost of DNNs.
However, the trade-off between the quantization bitwidth and final accuracy is complex and non-convex, which makes it difficult to be optimized directly. 
Minimizing direct quantization loss (DQL) of the coefficient data is an effective local optimization method, 
but previous works often neglect the accurate control of the DQL, resulting in a higher loss of the final DNN model accuracy.
In this paper, we propose a novel metric, called Vector Loss.
Using this new metric, we decompose the minimization of the DQL to two independent optimization processes, 
which significantly outperform the traditional iterative L2 loss minimization process in terms of effectiveness, quantization loss as well as final DNN accuracy.
We also develop a new DNN quantization solution called VecQ, which provides minimal direct quantization loss and achieve higher model accuracy.
In order to speed up the proposed quantization process during model training, we accelerate the quantization process with a parameterized probability estimation method and template-based derivation calculation.
We evaluate our proposed algorithm on MNIST, CIFAR, ImageNet, IMDB movie review and THUCNews text data sets with numerical DNN models. The results demonstrate that our proposed quantization solution is more accurate and effective than the state-of-the-art approaches yet with more flexible bitwidth support.
Moreover, the evaluation of our quantized models on Saliency Object Detection (SOD) tasks maintains comparable feature extraction quality with up to 16$\times$ weight size reduction. 
\end{abstract}
}
\maketitle
\IEEEdisplaynontitleabstractindextext
\IEEEpeerreviewmaketitle
\section{Introduction}\label{sec:introduction}

\IEEEPARstart{D}{eep} Neural Networks (DNNs) have been widely adopted in machine learning based applications \cite{simonyan2014vgg,he2016resnet}.
However, besides DNN training, DNN inference is also a computation-intensive task which affects the effectiveness of DNN based solutions~\cite{chen2019clouddnn,cong2019dac,zhang2018dnnbuilder}. 
Neural network quantization employs low precision and low bitwidth data instead of high precision data for the model execution. 
Compared to the DNNs with floating point with 32-bit width (FP32), the quantized model can achieve up to 32$\times$ compression rate with an extremely low-bitwidth quantization \cite{hubara2016binarized}.
The low-bitwidth processing, which reduces the cost of the inference by using less memory and reducing the complexity of the multiply-accumulate operation, improves the efficiency of the execution of the model significantly~\cite{zhang2018dnnbuilder,hybridfpga}.
    
However, lowering the bitwidth of the data
often brings accuracy degradation~\cite{cong2019dac,chen2019tdla,gysel2016hardware}. 
This requires the quantization solution to balance between computing efficiency and final model accuracy.
However, the quantitative trade-off is non-convex and hard to optimize 
-- \emph{the impact of the quantization to the final accuracy of the DNN models is hard to formulate}.

Previous methods neglect the quantitative analysis of the \textbf{D}irect \textbf{Q}uantization \textbf{L}oss (DQL) of the weight data and make the quantization decision empirically while directly evaluating the final model accuracy~\cite{hubara2016binarized,Binaryconnect,Ternaryconnect,zhou2016dorefa,jin2018sparse} thus only achieving unpredictable accuracy.

In order to achieve higher training accuracy,
finding an optimal quantization solution with minimal loss during the training of the learning kernels is effective and practical.
One way of finding a local optimal solution is to minimize the DQL of the weight data, which is widely used in the current quantization solutions~\cite{han2015deep, TWNs,ENN2017,TSQ2018,cheng2019uL2Q}.

\begin{figure}
    \centering
    \includegraphics[width=0.4\textwidth]{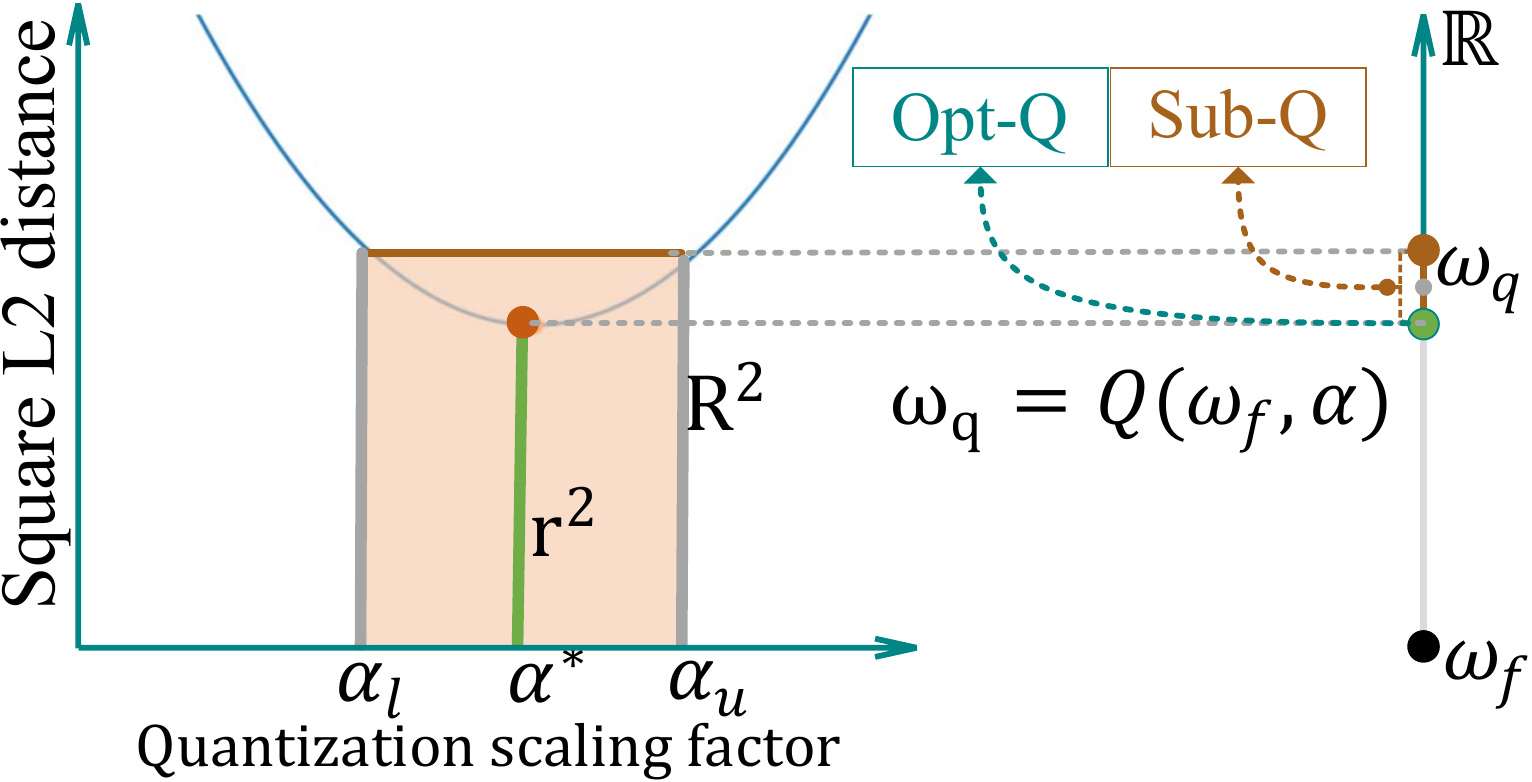}
    \vspace{-5pt}
    \caption{\label{fig:l2_distance_vecter_dql}
    Uncertainty of using L2 to evaluate the quantization loss. $\alpha^*$ is the optimal scaling factor for the quantization in the range of [$\alpha_l$, $\alpha_u$]. The optimal distance and achievable distance are denoted as $r^2$ and $R^2$.}
    \vspace{-15pt}
\end{figure}

As shown in \figref{fig:l2_distance_vecter_dql}, $w_f$ denotes the full-precision weight and $w_q$ is the value after quantization. 
Conventional quantization methods regard $w_f$ as a point (set as origin in \figref{fig:l2_distance_vecter_dql}) in Euclidean Space, and $w_q$ is a point which is close to $w_f$ in a discrete data space. The discrete data space contains a certain number of data points that can be represented by the selected bitwidth.
Therefore, the Square of Euclidean Distance (Square 2-norm or called L2 distance~\cite{zhu2016TTQ}), represented as $||w_f-w_q||_2^2$, between the original weight data and the quantized data is simply used as the loss of the quantization process, which is going to be reduced~\cite{TWNs,ENN2017,TSQ2018,cheng2019uL2Q}. 

Although the L2 based solutions are proven to be effective and provide good training results in terms of accuracy of the model and bitwidth of the weight data, such solutions still have some major issues.
(1) Traditional L2 based optimizations generally rely on iterative search methods and can get stuck at local optima. As shown in \figref{fig:l2_distance_vecter_dql},
the quantized results usually fall into the sub-optimal space (Sub-Q) instead of the optimal value (Opt-Q).
Even with an additional quantization scaling factor $\alpha$, which could help to reduce the differences between the original and quantized data, traditional methods still can not avoid considerable accuracy loss during the quantization process. 
(2) The process of L2 based quantization focuses on each of the individual weight data and neglects the distribution and correlations among these data points in a kernel or a layer.

To address the issues above, instead of directly minimizing L2 loss, we propose a more accurate quantization loss evaluation metric called \textit{Vector Loss}. 
Minimizing vector loss is much efficient than directly and iteratively minimizing L2, and results in higher final DNN accuracy;
we also propose an algorithm to guide the quantization of the weight data effectively.
We construct the weights into a vector $\boldsymbol{w_f}$ rather than scalar data,
to take advantage of the characteristic that the loss between vectors can be decomposed into orientation loss and modulus loss, which are independent of each other.
As a result, we are able to achieve the minimal loss of the weight quantization for DNN training.

In this paper, we will demonstrate that using vectorization loss as an optimization objective is better than directly optimizing the L2 distance of the weights before and after quantization.
Based on our proposed vectorized quantization loss measurement, we further propose a \textbf{Vec}torized \textbf{Q}uantization method (\textbf{VecQ}) to better explore the trade-off between computing efficiency and the accuracy loss of quantization.

In summary, our contributions are as follows:
\begin{itemize}
\item 
    We propose a new metric, \textit{Vector Loss}, as the loss function for DNN weight quantization, which can provide effective quantization solution compared to traditional methods.
\item 
    A new quantization training flow based on the vectorized quantization process is proposed, named VecQ, which 
    achieves better model accuracy for different bitwidth quantization target.
\item 
    Parametric estimation and computing template are proposed to reduce the cost of probability density estimation and derivative calculation of VecQ to speed up the quantization process in model training. 
\item 
    Extensive experiments show that VecQ achieves a lower accuracy degradation under the same training settings when compared to the state-of-the-art quantization methods in the image classification task with the same DNN models.
    The evaluations on Saliency Object Detection (SOD) task also show that our VecQ maintains comparable feature extraction quality with up to 16$\times$ weight size reduction.
\end{itemize}

This paper is structured as follows. Section~\ref{sec:relateworks} introduces the related works. In Section~\ref{sec:equdeduc}, 
the theoretical analysis of the effectiveness of vector loss compared to L2 loss is presented.
Section~\ref{sec:vector_quantization} presents the detailed approach of VecQ. Section \ref{sec:fast_quantization} proposes the fast solution for our VecQ quantization as well as the integration of VecQ into the DNN training flow. Section \ref{sec:experiments} presents the experimental evaluations and Section \ref{sec:conclusion} concludes the paper.
\section{Related works and Motivation}\label{sec:relateworks}
As an effective way to compress DNNs, many quantization methods have been explored~\cite{gysel2016hardware,hubara2016binarized,rastegari2016xnor,Binaryconnect,TWNs,zhu2016TTQ,alemdar2017ternary,Ternaryconnect,cheng2019uL2Q,wang2019haq,INQ2017,jin2018sparse,ghasemzadeh2018rebnet,TSQ2018,ENN2017,deng2018gxnor,chen2019tdla,QAT,TQT,han2015deep,yu2018gradiveq,zhou2016dorefa}.
These quantization methods can be roughly categorized into 3 different types based on their objective functions for the quantization process:
\begin{itemize}
    \item Methods based on heuristic guidance of the quantization, e.g., directly minimizing the final accuracy loss;
    \item Methods based on minimizing Euclidean Distance of weight data before and after quantization;
    \item Other methods such as training with discrete weights and teacher-student network.
\end{itemize}

In this section, we first introduce the existing related works based on their different categories and then present our motivation for vectorized quantization.

\vspace{-2pt}
\subsection{Heuristic guidance}
The heuristic methods usually directly evaluate the impact of the quantization on the final output accuracy.
They often
empirically iterate the training process to improve the final accuracy.
For example, the BNNs \cite{hubara2016binarized} proposed a binary network for fast network inference. 
It quantizes all the weights and activations in a network to 2 values, $\{-1,+1\}$, based on the sign of the data. 
Although it provides a DNN with 1-bit weights and activations, it is hard to converge without Batch Normalization layers~\cite{ioffe2015batch} and leads to a significant accuracy degradation when compared to full-precision networks.
The Binary Connect \cite{Binaryconnect} and Ternary Connect \cite{Ternaryconnect} sample the original weights into binary or ternary according to a sampling probability defined by the value of the weights (after scaling to [0,1]). 
All these works do not quantify the loss during the quantization, so that only the final accuracy is the guideline of the quantization.

Quantization methods in \cite{zhou2016dorefa,gysel2016hardware}
convert the full-precision weights to fixed-point representation by dropping the least significant bits without quantifying the impact.

INQ \cite{INQ2017} iteratively processes weight partition, quantization and re-training method until all the weights are quantized into powers-of-two or zeros.

STC \cite{jin2018sparse} introduces a ternary quantization which first scales the weights into the range of $[-1,1]$, and then quantizes all scaled weights into ternary by uniformly partitioning them. 
Thus, the values located in $[-1,-1/3]$ and $[1/3,1]$ are quantized to -1, 1 and the rest of them are set to 0.

TTQ \cite{zhu2016TTQ} introduces a ternary quantization which quantizes full-precision weights to ternary by a heuristic threshold but with two different scaling factors for positive and negative values, respectively. The scaling factors are optimized during the back propagation. 

The quantization method in~\cite{QAT} (denoted as QAT) employs the \emph{affine mapping} of integers to real values with two constant parameters: Scale and Zero-point. It first subtracts the Zero-point parameter from data (weights/activation), then divides the data by a scaling factor and obtains the quantized results with rounding operation and affine mapping. 
The approach of TQT~\cite{TQT} follows QAT but with the improvement of constraining the scale-factors into power-of-2 and relates them to trainable thresholds.

\vspace{-4pt}
\subsection{Optimizing Euclidean Distance}
In order to provide better accuracy control, reducing the Euclidean Distance of the data before and after quantization becomes a popular solution.

Xnor-Net \cite{rastegari2016xnor} adds a scaling factor on the basis of BNNs \cite{hubara2016binarized} and calculates the optimal scaling factor to minimize the distance of the weights before and after quantization. The scaling factor boosts the convergence of the model and improves the final accuracy. 
The following residual quantization method in~\cite{ghasemzadeh2018rebnet}
adopts Xnor-Net \cite{rastegari2016xnor} to further compensate the errors produced by single binary quantization to improve the accuracy of the quantized model.

TWN \cite{TWNs} proposes an additional threshold factor together with the scaling factor for ternary quantization. 
The optimal parameters (scaling factor and threshold factor) are still based on the optimization of the Euclidean distance of weights before and after quantization. TWN achieves better final accuracy than Xnor-Net and BNNs. 

Extremely low bit method (ENN) proposed in~\cite{ENN2017} quantizes the weights into the exponential values of 2 by iteratively optimizing the L2 distance of the weights before and after quantization.

TSQ \cite{TSQ2018} presents a two-step quantization method, which first quantizes the activation to low-bit values, and then fixes it and quantizes the weights into ternary. 
TSQ employs scaling factor for each of the kernel, resulting in a limited model size reduction.

$\mu$L2Q \cite{cheng2019uL2Q} first shifts the weights of a layer to a standard normal distribution with a shifting parameter and then employs a linear quantization for the data.
The uniform parameter considers the distribution of the weight data, which provides better loss control compared to simply optimizing the Euclidean Distance during the quantization.

Several other works \cite{han2015deep,yu2018gradiveq} 
adopt k-means with irregular non-linear quantization. Although the values are clustered before quantization, the final results are still obtained with the optimization of the Euclidean distance between the original values and the quantized ones. 

\vspace{-2pt}
\subsection{Other works}
Besides the heuristic and Euclidean Distance approaches, there are still many other works focusing on low-precision DNN training.

GXNOR-Net \cite{deng2018gxnor} utilizes the discrete weights during training instead of the full-precision weights. 
It regards the discrete values as states and projects the gradients in backward propagation as the transition of the probabilities to update the weights directly, hence, providing a network with ternary weights.

T-DLA \cite{chen2019tdla} quantizes the scaling factor of ternary weights and full-precision activation into fixed-point numbers and constrains the quantization loss of activation values by adopting infinite norms. 
Compared with \cite{gysel2016hardware,zhou2016dorefa},
it shifts the available bitwidth to the most effective data portion to make full use of the targeted bitwidth.

In TNN \cite{alemdar2017ternary}, the authors design a method using ternary student network, which has the same network architecture as the full-precision teacher network, aiming to predict the output of the teacher network without training on the original datasets.

In HAQ \cite{wang2019haq}, the authors proposed a range parameter --- all weights out of the range are truncated and the weights within the range are linearly mapped to discrete values. The optimal range parameter was obtained by solving the KL-divergence of the weights during the quantization.

However, comparing to the heuristic guidance and Euclidean Distance based methods, the approaches above either focus on a specific bitwidth or perform worse in terms of the accuracy of the trained DNNs.

\vspace{-3pt}
\subsection{Motivation of the VecQ Method}
For the sake of simplicity, in this paper, we will use L2 distance to represent the squared L2 distance.
We have witnessed the effectiveness of the L2 distance-based methods among all the existing approaches.
However, as explained in the introduction, there are still two defects that lead to inaccurate DQL measurement.

The first defect is that the traditional way of using L2 distance as the loss function for optimization
usually cannot be solved accurately and efficiently~\cite{TWNs,cheng2019uL2Q,TSQ2018,han2015deep,yu2018gradiveq}, even with an additional scaling factor $\alpha$ to scale the data into proper range.
As shown in \figref{fig:l2_distance_vecter_dql}, the quantization function with the additional scaling factor $\alpha$ to improve the accuracy~\cite{cheng2019uL2Q, TWNs} is denoted as $w_q=Q(w_f,\alpha)$;
the L2 distance curve between $w_f$ and $w_q$ with the change of $\alpha$ is drawn in blue.
It has a theoretical optimal solution when $\alpha = \alpha^*$ with a L2 distance of $r^2$, shown as the green dot.
However, only the solutions $\alpha \in [\alpha_l,\alpha_u]$ with the L2 distance ranging in $[r^2, R^2]$ could be obtained due to the lack of solvable expressions~\cite{TWNs,cheng2019uL2Q}, leading to an inefficient and inaccurate quantization result.
Additionally, even the methods involving k-means for clustering of the weights still fall into the sub-optimal solution space~\cite{han2015deep,yu2018gradiveq}.
Their corresponding approximated quantized weights are located in the \textbf{Sub-Q} space colored with orange.

The second defect of traditional methods is they
neglect the correlation of the weights within the same kernel or layer, but only focus on the difference between single values. Even with the k-means based solutions, the distribution of the weights in the same layer is ignored in the quantization process. However, the consideration of the distribution of the weight data is proven to be effective for the accuracy control in the existing approaches~\cite{cheng2019uL2Q, chen2019tdla}.

We discover that when we represent quantization loss of the weight for a kernel or a layer using vector distance instead of L2 distance, it will intrinsically solve the two problems mentioned above.
We focus on two attributes of a vector, orientation and modulus.
Meanwhile,
we define a Quantization Angle that represents the intersection angle between the original weight vector and the quantized vector.
As a result, the vector distance between the two is naturally determined by the Quantization Angle and their modulus.
Therefore, in this work, we evaluate DQL with Vector Loss by leveraging the vector distance, which involves both quantization angle and vector modulus.
Note the orientation of the vector is related to quantization angle. When 
the quantization angle is 0, the orientations of the two intersecting vectors are the same. 
When the angle is not 0, the orientations of the two vectors are different. For the sake of simplicity, 
in the rest of the paper, we will use orientation loss and modulus loss to represent the Vector Loss. 
To the best of our knowledge, there is no previous work that leverages the vector loss for DNN weight quantization. 

In this work, we demonstrate that the vector loss can provide effective quantization solution and hence achieve a smaller DQL for the weight data quantization during the model training, which helps for achieving a higher model accuracy.
Based on this, we propose VecQ, which carries out the quantization process based on vector loss. 
We also propose a fast parameter estimation method and a computation template to speed up our vectorized quantization process for easier deployment of our solution.
\section{Vector Loss Versus L2 loss}\label{sec:equdeduc}
\begin{figure}
    \centering
    \includegraphics[width=0.98\columnwidth]{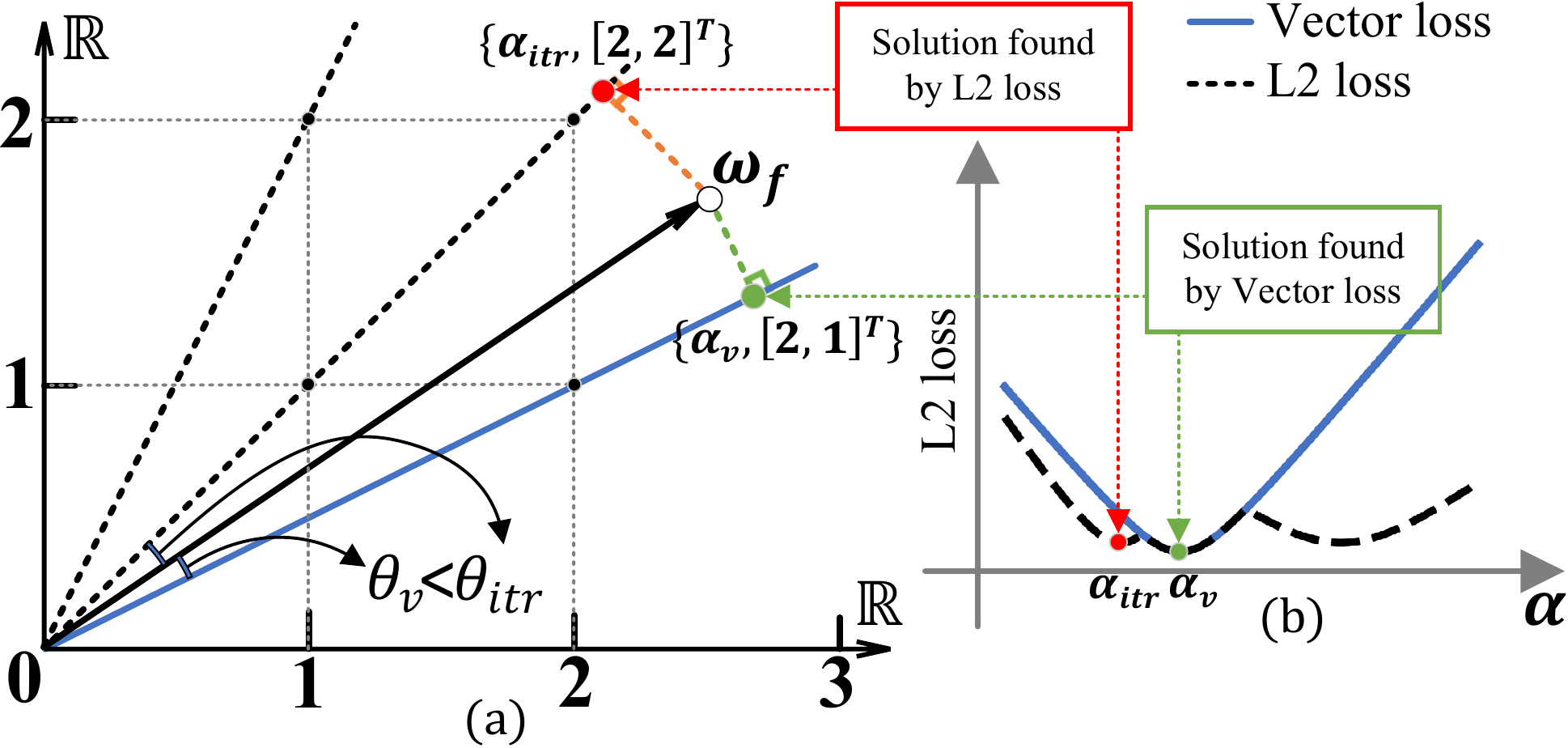}
    \vspace{-5pt}
    \caption{The solutions based on L2 loss and Vector loss. (a) Solutions of different methods; (b) L2 Loss with different methods and $\alpha$ values.}
    \label{fig:l2_vs_vecq}
    \vspace{-15pt}
\end{figure}

Before introducing our vectorized quantization method, we first explain the effectiveness of loss control with vector loss using two data points as an example for simplicity.
Assume a DNN layer with only two weights, denoted as \{$w_{f1},w_{f2}$\} whose values are \{2.5, 1.75\}. The weights will be quantized into $k$ bits.
The quantization loss based on L2 distance is denoted as $J_{l2}$ and the quantization solution set is expressed in the format of \{$\alpha, (v_1, v_2)$\}, where $\alpha$ is the floating point scaling factor and $v_1, v_2$ are the discrete values in the quantization set $\{v_1, v_2\} \in Q = \{-2^{k-1},-2^{k-1}+1,\cdots,-1,0,1,\cdots,2^{k-1}-1\}$, then we get
\begin{equation}
    J_{l2} = (w_{f1} - \alpha v_{1})^2 + (w_{f2} - \alpha v_{2})^2
\end{equation}
Let $\boldsymbol{w_f}=[w_{f1},w_{f2}]^T$ be a vector from the origin point $(0,0)$, and its quantized value is $\boldsymbol{w_q}=\alpha\boldsymbol v$ and $\boldsymbol v=[v_{1},v_{2}]^T$. 
$J_{l2}$ could also be represented as the squared modulus of the distance between vector $\boldsymbol{w_f}$ and $\boldsymbol{w_{q}}$.
The $J_{l2}$ is calculated as:
\begin{equation}
\vspace{-2mm}
    J_{l2}=||\boldsymbol w_f-\alpha\cdot \boldsymbol v||_2^2, ~ (v_{i} \in Q, i \in [1, 2^k])
\end{equation}

As shown in \figref{fig:l2_vs_vecq} (a), there are only two dimensions in the solution space, each representing one weight. Each dimension contains $2^{k}$ possible values.
The values of the possible solutions are located on the black dotted line in the Fig.~\ref{fig:l2_vs_vecq} due to the full precision scaling factor $\alpha$.
The quantization angle between $\boldsymbol{w_f}$ and the quantized version is denoted as $\theta$.

However, due to the non-convex characteristic of optimizing L2 loss under the $k$-bit constraint~\cite{ENN2017}, the result based on the iterative method may be found as the red point in the figure with the solution of \{$\alpha_{itr}, \boldsymbol{v_{itr}}\} = \{1.0625, [2, 2]^T$\} and an angle of $\theta_{itr}$, which is the first sub-optimal solution point on the curve of the L2 loss vs $\alpha$ values in~\figref{fig:l2_vs_vecq} (b), ignoring the solution with lower loss which is the second extreme value on the curve.

Therefore, instead of directly using L2 distance as the quantization loss function, 
we use Vector Loss, denoted as $J_v$, to measure the difference between vectors: the original weight vector $\boldsymbol{w_f}$ and the quantized weight vector $\boldsymbol w_q$, to obtain the quantization solution.
We define the vector loss $J_v = J(\boldsymbol w_{f}, \boldsymbol w_{q})$ as a composition of two independent losses, orientation loss $J_o$ and modulus loss $J_m$, and 
\begin{equation}\label{eq:jvdef}
J_v = J_o + J_m.
\end{equation}
$J_o$ and $J_m$ are computed as:
\begin{equation}
\begin{split}
    J_o&=1-\cos\theta,~(\cos\theta=\frac{\alpha\boldsymbol v}{|\alpha\boldsymbol v|}\frac{\boldsymbol w_f}{|\boldsymbol w_f|}) \\
    &=1-\boldsymbol{e_{v}} \boldsymbol{e_{w_f}}\\
    &=1-\sum_{i=1}^{d}(e_{v_i}e_{w_{fi}}) \\
    J_m&=||\boldsymbol w_f-\alpha\boldsymbol v||_2^2
\end{split}
\label{eq:decomposition}
\end{equation}
where $\boldsymbol{e_v}$ and $\boldsymbol{e_{w_f}}$ represent the unit vector for $\boldsymbol{v}$ and $\boldsymbol{w_f}$.
$\boldsymbol{w_f}$ is a weight vector of a layer of a DNN containing $d$ weights.

Given the definition of $J_v$, we now discuss the effectiveness of minimizing $J_v$.


First, according to Eq.~\ref{eq:decomposition}, the optimization of the orientation loss is only defined by the unit vectors $\boldsymbol{e_{v}}$ and $\boldsymbol{e_{w_{f}}}$, where $\alpha$ value is not affected.
Second, when the $\boldsymbol{e_{v}}$ is determined, the optimization of the modulus loss is to find the optimal value of $\alpha$. 
Therefore, the optimization of $J_{o}$ and $J_{m}$ can be achieved independently.
If both $\theta$ and $\alpha$ are continuous, both orientation and modulus loss minimizations are convex and can be solved with the optimal solutions. However, due to the integer constraints coming from the quantized bitwidth, we need to take the floating-point results and find the corresponding integer values. As a result, the actual achievable $\theta_{v}$ of our solution can be different from the optimal value for it. Thus, our solution represents an approximated solution of the optimal solution.


Furthermore, denote the quantization result of optimizing $J_{v}$ as $\boldsymbol w_{q}^{v}$, we have:
\vspace{-4pt}
\begin{equation}\label{eq:theta}
    \begin{split}
    |\boldsymbol w_q^v|&= |\boldsymbol w_f|\cos\theta_v\\
    |\boldsymbol w_q^{itr}|&=|\boldsymbol w_f|\cos\theta_{itr}
    \end{split}
\end{equation}

Typically, our approximated result can satisfy ${\theta_{v}}\le{\theta_{itr}}$, then we could easily achieve:

\begin{equation}
\label{eq:loss_comparison}
    \begin{split}
    J_{l2}=&|\boldsymbol{w_f}-\boldsymbol{w_q^v}|^2\\
    =&|\boldsymbol{w_f}|^2+|\boldsymbol{w_q^v}|^2-2|\boldsymbol{w_f}||\boldsymbol{w_q^v}|\cos{\theta_{v}}\\
    =&|\boldsymbol{w_f}|^2-|\boldsymbol{w_f}|^2\cos^2{\theta_{v}}\\
     \le&|\boldsymbol{w_f}|^2-|\boldsymbol{w_f}|^2\cos^2{\theta_{itr}}\\
    =&|\boldsymbol{w_f}-\boldsymbol{w_q^{itr}}|^2
    \end{split}
\end{equation}
which shows that with vector loss, as long as we optimize the orientation loss to find a better quantization angle, we could achieve an effective solution for weight quantization.
Guided by the observation above, we have the final optimal solution for the example points located on the blue solid line in the Fig.~\ref{fig:l2_vs_vecq} as \{$\alpha_{v}, \boldsymbol{v}_{v}\} = \{1.35,[2,1]^T$\}, which provides a smaller DQL.

Base on our proposed vector loss metric, we will discuss our algorithms to obtain the vector loss based quantization solution as well as the methods to speed up the algorithms for practical purpose.
\section{Vectorized Quantization}\label{sec:vector_quantization}

VecQ is designed to follow the theoretical guideline we developed
in \secref{sec:equdeduc}.
First of all, the vectorization of the weights is introduced.
Then the adoption of the vector loss in VecQ is explained in detail.
Finally, the process of VecQ quantization with two critical stages is presented.

\subsection{Vectorization of weights}
For the weight set $\mathrm{W_f}(l)$ of layer $l$, 
we flatten and reshape them as a $N \times M \times K^2$-dimension vector $\boldsymbol{w_f}(l)$.
$N$, $M$~indicate the number of input channel and output channel for this layer, and $K$ indicates the size of the kernel for this layer.
For simplicity, we use $\boldsymbol{w_f}$ to represent the weight vector of a certain layer before quantization.

\subsection{Loss function definition}\label{subsec:loss}
We use the vectorized loss instead of Euclidean distance during the quantization process.
Since solving the orientation and modulus loss independently could achieve the optimized solution for each of them,
we illustrate the quantization loss as is defined in Equ.~\ref{eq:jvdef}:
\begin{equation}
    \label{eq:quantization_loss}
    J(\boldsymbol{w_f},\boldsymbol{w_q})=J_{o}(\boldsymbol{w_f},\boldsymbol{w_q})+J_{m}(\boldsymbol{w_f},\boldsymbol{w_q})
\end{equation}
to provide the constraint during the quantization process.

\subsection{Overall process}

\begin{figure}[ht]
    \centering
    \includegraphics[width=0.4\textwidth]{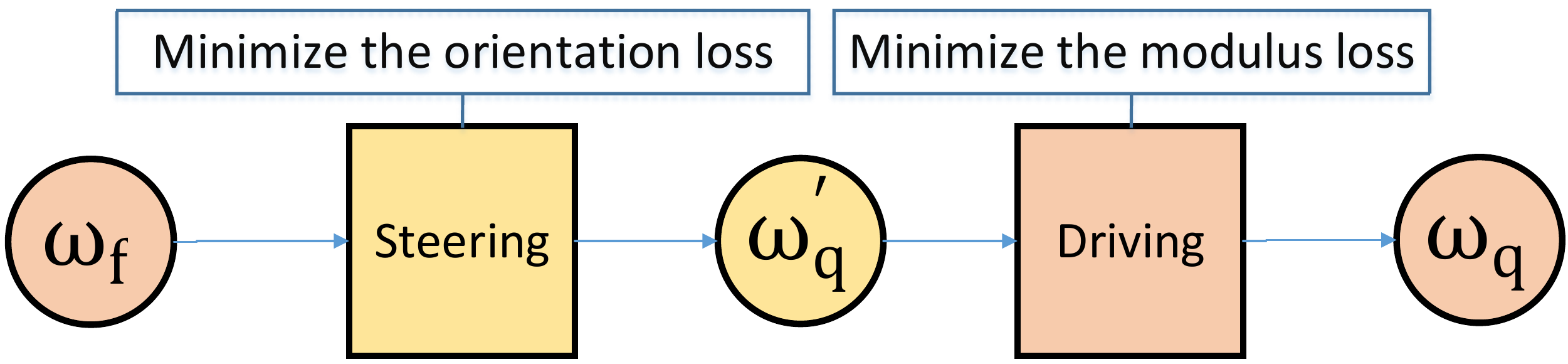}
    \vspace{-6pt}
    \caption{The overall flow of quantization process, including both steering and driving stage.}
    \label{fig:quantization_process}
    \vspace{-4pt}
\end{figure}

According to our analysis in Section~\ref{sec:equdeduc}, the orientation loss $J_{o}$ indicates the optimized quantization angle and the modulus loss $J_{m}$ indicates the optimized scale at this angle.
Therefore, our quantization takes two stages to minimize the two losses independently,
which are defined as \textbf{steering stage} and \textbf{driving stage} as shown in \figref{fig:quantization_process}. 
In the steering stage, we adjust the orientation of the weight vector to minimize the orientation loss.
Then, we fix the orientation and only scale the modulus of the vector at the driving stage to minimize the modulus loss.

Let $\boldsymbol{w_f} \in \mathbb{R}^{N\times M \times K^2}$ be the weight vector of the layer of a DNN in the real space and the $\boldsymbol{w_q} \in \mathbb{Q}^{N\times M \times K^2}$ be the quantized weight vector in the uniformly discrete subspace. 
First, steering $\boldsymbol{w_f}$ to $\boldsymbol{w_q'}$:
\begin{equation}
    \boldsymbol{w_q'}=steer(\boldsymbol{w_f})
\end{equation}
Where $\boldsymbol{w_q'}$ is an orientation vector that disregards the modulus of the vector and only focuses on the orientation.
Second, along with the determined orientation vector $\boldsymbol{w_q'}$, we search the position of the modulus and "drive" to the best position with minimum modulus loss.
The quantized vector $\boldsymbol{w_q}$ is achieved by driving the $\boldsymbol{w_q'}$.
\begin{equation}
    \boldsymbol{w_q}=drive(\boldsymbol{w_q'})
\end{equation}

The complete quantization process is the combination of the two stages. The final target is reducing the loss between the original weight $\boldsymbol{w_f}$ and the quantized results $\boldsymbol{w_q}$.
The entire quantization process is represented as
\begin{equation}
    \label{eq:vecq}
    \boldsymbol{w_q} = Q(\boldsymbol{w_f})=drive(steer(\boldsymbol{w_f}))
\end{equation}

\subsection{Steering stage}\label{sec:steering}

The purpose of the steering stage is to search for an optimized orientation vector, which has the least orientation loss with $\boldsymbol{w_f}$ to minimize the $J_o$.
\begin{figure}[ht]
    \centering
    \includegraphics[width=0.4\textwidth]{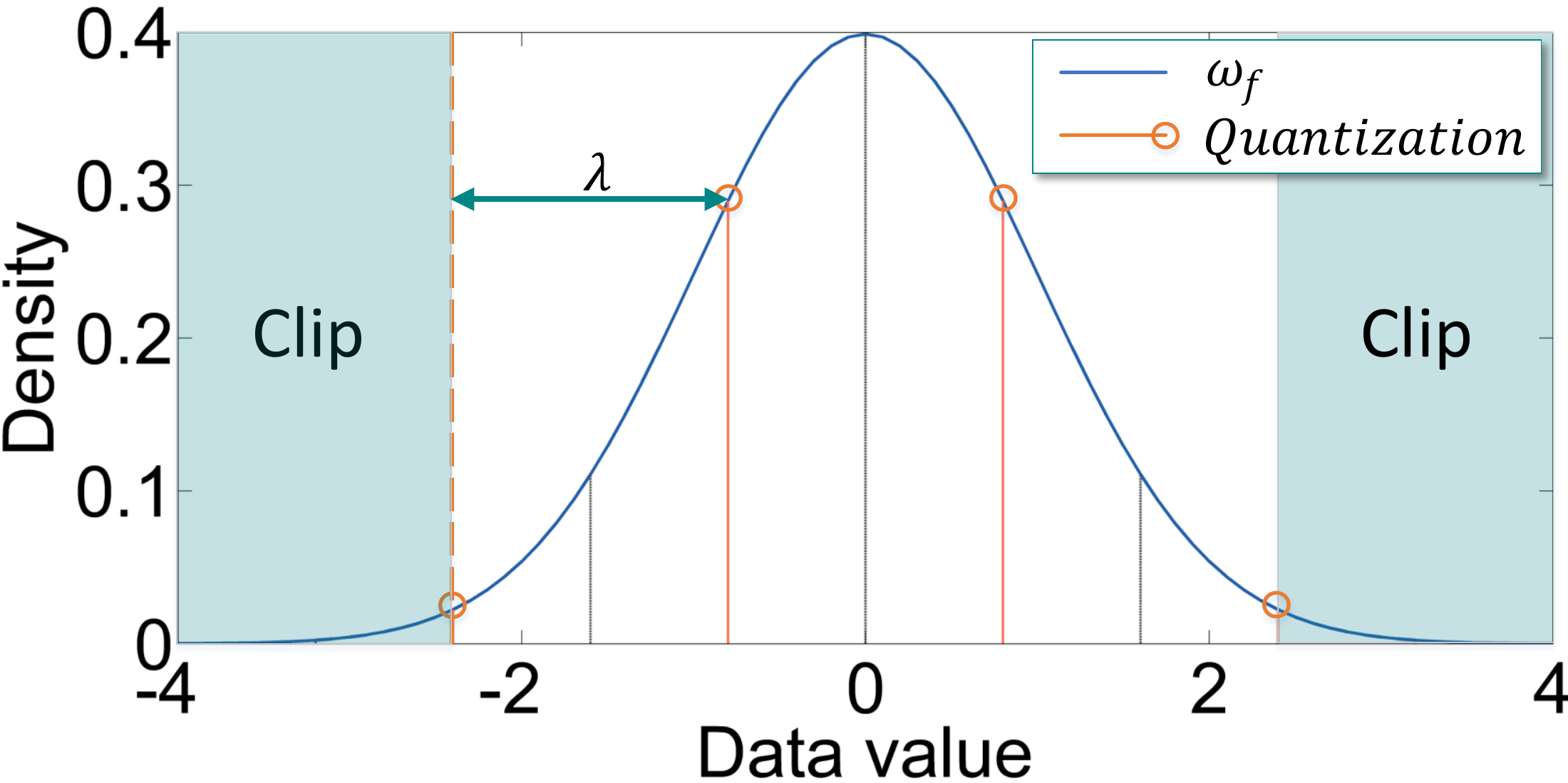}
    \vspace{-2pt}
    \caption{Linear quantization with interval $\lambda$.}
    \vspace{-4pt}
    \label{fig:uniform_quantization}
\end{figure}

As shown in \figref{fig:uniform_quantization}, $\boldsymbol{w_f}$ is the  weight in floating point representation and it would be quantized into $k$-bit representation. It means, there are total $2^k$ values that can be used to represent the values in $\boldsymbol{w_f}$, where each of them is denoted as $q_{i}, i \in[1,2^k]$. 
We adopt linear quantization method, where an interval $\lambda = q_{i} - q_{i-1}$ is used to represent the distance between two quantized values.
\begin{equation}
    \label{eq:steer}
    steer(\boldsymbol{w_f},\lambda,k)=Clip\{\lfloor\frac{\boldsymbol{w_f}}{\lambda}-0.5\rceil,-2^{k-1},2^{k-1}-1\}+0.5
\end{equation}

The vector with floating data $\boldsymbol{w_f}$ is quantized to a vector with discrete data by an rounding ($\lfloor\cdot\rceil$) operation for each of the values in the vector.
The data are limited to the range of $[-2^{k-1},2^{k-1}-1]$ by extended clip ($Clip()$) operation.
The subtraction of $0.5$ is used to avoid aggregation at 0 position and guarantees the maximum number of rounding values to be $2^k$.

Given a $k$ for the number of bits as the quantization target, the intermediate quantized weight is
\begin{equation}
    \boldsymbol{w_q'}=steer(\boldsymbol{w_f},\lambda,k) 
\end{equation}
which has the minimized orientation loss with the $\lambda$ as an interval parameter. 
When $k$ is fixed, $\lambda$ decides the orientation loss between $\boldsymbol{w_f}$ and $\boldsymbol{w_q'}$. 
In order to minimize the loss, we only need to find the optimal $\lambda$:
\begin{equation}
    \label{eq:optimal_lambda}
    \lambda^*=\arg\underset{\lambda}{\min}(J_{o}(\boldsymbol{w_f},\boldsymbol{w_q'}))
\end{equation}

Finding the optimal $\lambda$ requires several processes with high computational complexities; the detailed processes and the corresponding fast solution is presented in Section~\ref{sec:fast_quantization}.

\subsection{Driving stage}

In the driving stage, we minimize the modulus loss $J_m$ between the orientation vector $\boldsymbol{w_q'}$ obtained from the steering stage and the original weight vector $\boldsymbol{w_f}$.
Since we focus on the modulus in this stage, only the scaling factor $\alpha$ is involved.
\vspace{-3pt}
\begin{equation}
\label{eq:drive}
    drive(\boldsymbol{w_q'})=\alpha \boldsymbol{w_q'}
\end{equation}

Here we only need to find the optimized $\alpha$ to minimize the modulus loss.
\begin{equation}
    \alpha^*=\arg\underset{\alpha}{\min}(J_m(\boldsymbol{w_f},\boldsymbol{w_q}))
\end{equation}
where $\alpha^*$ can be easily obtained by finding the extreme of $J_m$ with
\begin{equation}
    \frac{\mathrm{d}J_m}{\mathrm{d} \alpha}=-2{\boldsymbol{w_{q}'}}^T(\boldsymbol{w_f}-\alpha\boldsymbol{w_q'})=0
\end{equation}
and the value of
$\alpha^*=\frac{{\boldsymbol{w_{q}'}}^T\boldsymbol{w_f}}{{\boldsymbol{w_{q}'}}^T{\boldsymbol{w_{q}'}}}$.

Finally, with the two stages above, the quantized value of the $\boldsymbol{w_f}$ is determined by $\lambda$ and $\alpha$:
\vspace{-5mm}
\begin{equation}
    \boldsymbol{w_q} = Q(\boldsymbol{w_f}), (\lambda \rightarrow \lambda^*, \alpha^* =\frac{{\boldsymbol{w_{q}'}}^T\boldsymbol{w_f}}{{\boldsymbol{w_{q}'}}^T{\boldsymbol{w_{q}'}}})
\end{equation}
\section{Fast Quantization}\label{sec:fast_quantization}

\begin{figure}[ht]
    \centering
    \includegraphics[width=0.45\textwidth]{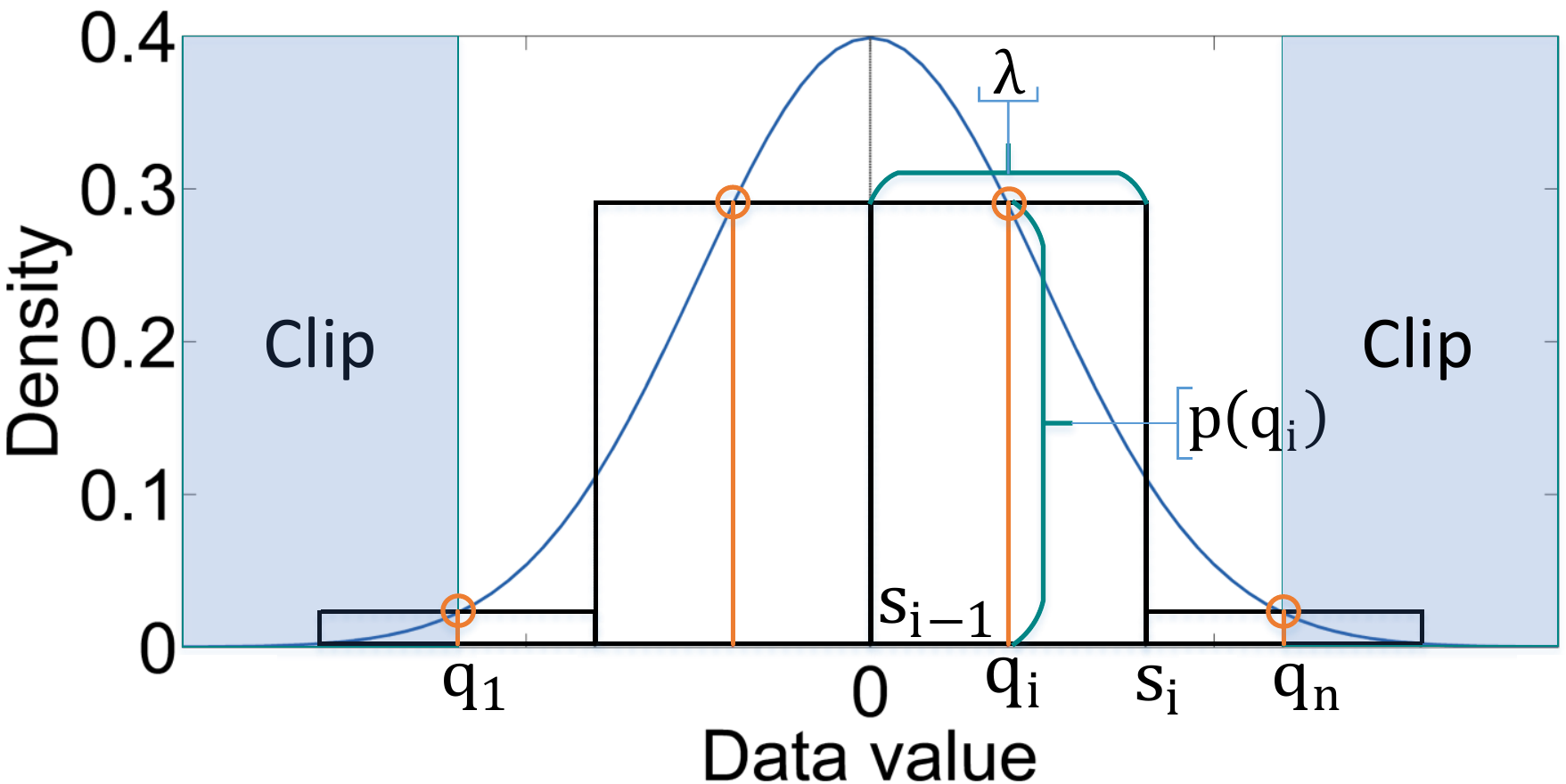}
    \vspace{-5pt}
    \caption{The illustration of quantization regions and symbols.}
    \label{fig:orientation_loss_computing}
    \vspace{-5pt}
\end{figure}

With the proposed quantization method in \secref{sec:vector_quantization}, the minimum quantization loss is achieved when the optimal $\lambda$ in \eqnref{eq:optimal_lambda}~is found.
However, as one of the most critical processes in the model training, the computational overhead of quantization leads to inefficient training of the model.
In order to address this issue, in this section, we first analyze the computational complexity to calculate the value of $\lambda$.
Then, we propose a fast solution based on our proposed fast probability estimation and computation template.
In the end, the detailed implementation of our quantization solution together with the fast solver is integrated into our training flow.

\subsection{Analysis of the optimal $\lambda$}

The most computational intensive process in our quantization is the steering stage, specifically, the process to solve \eqnref{eq:optimal_lambda}.
However, \eqnref{eq:optimal_lambda}~can not be solved directly due to the clipping and rounding operations.
Instead of directly using the values in $\boldsymbol{w_f}$, we involve the distribution of the values in $\boldsymbol{w_f}$ to support a more general computation of the $J_o$.
The probability density of the value $t$ in $\boldsymbol{w_f}$ is noted as $p(t)$.

According to the steering method in \eqnref{eq:steer}, each value $t \in \boldsymbol{w_f}$ is projected to a value $q \in \boldsymbol{w_q'}$;
the $q$ values are linearly distributed into $n~(n=2^k)$ with a uniform distance defined by interval $\lambda$.
As shown in \figref{fig:orientation_loss_computing}, the light blue curve is the distribution of values in $\boldsymbol{w_f}$ and orange dots are the values after quantization, represented as $q\in \boldsymbol{w_q'}$. 
$p(q_i)$ indicates the probability of $q_i$ in $\boldsymbol{w_f}$.

Specifically, the data within range $(s_{i-1},s_i]$ ~($s_i-s_{i-1}=\lambda, i=1,\cdots,n-1$) is replaced by the single value $q_i$ and the data out of the range $[q_1,q_n]$ are forced to be truncated and set to the nearest $q_i$.
As given in \eqnref{eq:steer}, $q_1=-2^{k-1}$ and $q_n=2^{k-1}-1$.

We set $s_0=-\infty$ and $s_n=\infty$ to ease the formulation.
Based on the distribution of data in $\boldsymbol{w_f}$, the expanding terms of $J_o(\boldsymbol{w_f},\boldsymbol{w_q'})$ in \eqnref{eq:optimal_lambda}~can be obtained as follows,
\begin{align}
\label{eq:terms_of_orientation_loss}
    \begin{cases}
    |\boldsymbol{w_f}|=\sqrt{\int_{-\infty}^{\infty}t^2p(t)\,\mathrm{d}t}\\
    |\boldsymbol{w_q'}|=\sqrt{\sum_{i=1}^n(\int_{s_{i-1}}^{s_i}q_i^2p(t)\,\mathrm{d}t)}\\
    \boldsymbol{w_f}\boldsymbol{w_q'}=\sum_{i=1}^n(\int_{s_{i-1}}^{s_i}tq_ip(t)\,\mathrm{d}t)
    \end{cases}
\end{align}
The $J_o(\boldsymbol{w_f},\boldsymbol{w_q'})$ is represented as
\begin{equation}
    \label{eq:expanding_formula_for_optimal_lambda}
    J_o(\boldsymbol{w_f},\boldsymbol{w_q'})=1-\frac{\sum_{i=1}^n(\int_{s_{i-1}}^{s_i}tq_ip(t)\,\mathrm{d}t)}{\sqrt{\int_{-\infty}^{\infty}t^2p(t)\,\mathrm{d}t}\cdot \sqrt{\sum_{i=1}^n(\int_{s_{i-1}}^{s_i}q_i^2p(t)\,\mathrm{d}t)}}
\end{equation}
Since the linear quantization is adopted with the fixed interval of $\lambda$, $q_i$ and $s_i$ can be easily derived by the following equations,
\begin{equation}
    q_i=(i-\frac{n}{2}-0.5)\lambda, (i=1,\cdots,n)
\end{equation}
\begin{equation}
 s_i=
 \begin{cases}
 -\infty,~i=0\\
 (i-\frac{n}{2})\lambda,~i=1,\cdots,n-1\\
 \infty,~i=n
 \end{cases}
\end{equation} 
So $J_o(\boldsymbol{w_f},\boldsymbol{w_q'})$ is only related to $\lambda$ and \eqnref{eq:optimal_lambda}~can be solved by solving
\begin{equation}
    \label{eq:partial_orientation}
    \frac{\mathbf{d} J_{o}(\boldsymbol{w_f},\boldsymbol{w_q'})}{\mathbf{d} \lambda}=0
\end{equation}

Concluding from the discussion above, for each $\boldsymbol{w_f}$, three steps are necessary: 
\begin{itemize}
    \item 
    Estimating probability density of the values in $\boldsymbol{w_f}$.
    \item 
    Solving \eqnref{eq:partial_orientation}~and getting the optimal $\lambda$.
    \item Using the optimal $\lambda$ to obtain the final quantization results.
\end{itemize}

However, the first two steps (probability density estimation and derivative calculation) are complex and costly in terms of CPU/GPU time and operations, which limit the training speed.

\subsection{Fast solver}
For the computation-intensive probability density estimation and derivative calculation, we propose two methods to speed up the processes, which are fast parametric estimation and computing template, respectively.

\subsubsection{Fast probability estimation}
There are two methods for probability density estimation: parametric estimation and non-parametric estimation.

Non-parametric estimation is usually used for fitting the distribution of data without prior knowledge.
It requires all the density and probability of the data to be estimated individually, which will lead to a huge computational overhead.

We take the widely adopted non-parametric estimation method, Kernel Density Estimation (KED) as an example, to illustrate the complexity of non-parametric estimation.
\begin{equation}
    p(t)=\frac{1}{nh}\sum_{i=1}^{n}K(\frac{t_i-t}{h})
\end{equation}
Here $p(t)$ is the probability density of $t$.
$n$ is the number of the samples and $K(x)$ is the non-negative kernel function that satisfies $K(x)\ge 0$ and $\int K(x)\mathrm{d}x=1$.
$h$ is the smoothing parameter.
The time complexity of computing all probability densities $p(x)$ is $O(n^2)$ and the space complexity is $O(n)$ because all the probability densities need to be computed and stored.

Parametric estimation is used for the data which have a known distribution and only computes some parameters of the distributions instead.
It could be processed fast with the proper assumption of the distribution.
Thus, we adopt a parametric estimation method in our solution.

There is prior knowledge of the weights of the layers of DNNs, 
which assumes that they are obeying normal distribution with the mean value of 0 so that the training could be conducted easily and the model could provide better generalization ability~\cite{cheng2019uL2Q,TSQ2018}:
\begin{equation}
\label{eq:normal_assumption}
    \boldsymbol{w_f} \sim \mathcal{N}(0,\sigma^2)
\end{equation}
$\sigma$ is the standard derivation of $\boldsymbol{w_f}$.
Based on this prior knowledge, we can use parametric estimation to estimate the probability density of $\boldsymbol{w_f}$ and the only parameter that is needed to be computed during the training is $\sigma$. The effectiveness of this prior distribution is also proven by the final accuracy we obtain in the evaluations.

With \eqnref{eq:normal_assumption}, we have
\begin{equation}
p(t)=\frac{1}{\sigma\sqrt{2\pi}} \, \exp \left( -\frac{t^2}{2\sigma^2} \right)
\end{equation}
Therefore, parametric estimation only requires the standard deviation $\sigma$, which could be calculated with
\begin{equation}
    \sigma^2=E(\boldsymbol{w_f}^2)-E(\boldsymbol{w_f})^2
\end{equation}
Here $E(\cdot)$ computes the expectation. Hence, the time complexity of computing $\sigma$ is reduced to $O(n)$ and the space complexity is reduced to $O(1)$.

\subsubsection{Computing template}

After reducing the complexity of computing the probability, finding optimal $\lambda$ is still complex and time-consuming.
In order to improve the computing efficiency of this step, we propose a computing template-based method.

Since the weights of a layer $\boldsymbol{w_f}$ obey normal distribution, $\mathcal{N}(0,\sigma^2)$, they can be transformed to the standard normal distribution:
\begin{equation}
    \boldsymbol{\varphi}=\frac{\boldsymbol{w_f}}{\sigma} \sim \mathcal{N}(0,1)
\end{equation}
Then, we could use $\boldsymbol{\varphi}$ to compute $J_{o}(\boldsymbol{\varphi},\boldsymbol{w_q'})$ instead of using $\boldsymbol{w_f}$, because of:
\begin{align}
    J_{o}(\boldsymbol{\varphi},\boldsymbol{w_q'})
    &=1-\frac{\boldsymbol{\varphi} \boldsymbol{w_q'}}{|\boldsymbol{\varphi}||\boldsymbol{w_q'}|}\nonumber\\
    &=1-\frac{(\boldsymbol{w_f}/\sigma) \boldsymbol{w_q'}}{|\boldsymbol{w_f}/\sigma||\boldsymbol{w_q'}|}\nonumber\\
    &=J_{o}(\boldsymbol{w_f},\boldsymbol{w_q'})
\end{align}
Here, $\varphi$ is the computing template for $\boldsymbol{w_f}$, because it has the same orientation loss with $\boldsymbol{w_q'}$ as $\boldsymbol{w_f}$.
By choosing this computing template, solving \eqnref{eq:partial_orientation}~is equivalent to solve the substitute equation \eqnref{eq:template_optimal_lambda}.
\begin{equation}
    \label{eq:template_optimal_lambda}
    \lambda^*=\arg\underset{\lambda}{\min}(J_{o}(\boldsymbol{\varphi},\boldsymbol{w_q'}))
\end{equation}
\begin{figure}[ht]
\vspace{-4pt}
    \centering
    \flushright
    \begin{overpic}[width=0.4\textwidth]{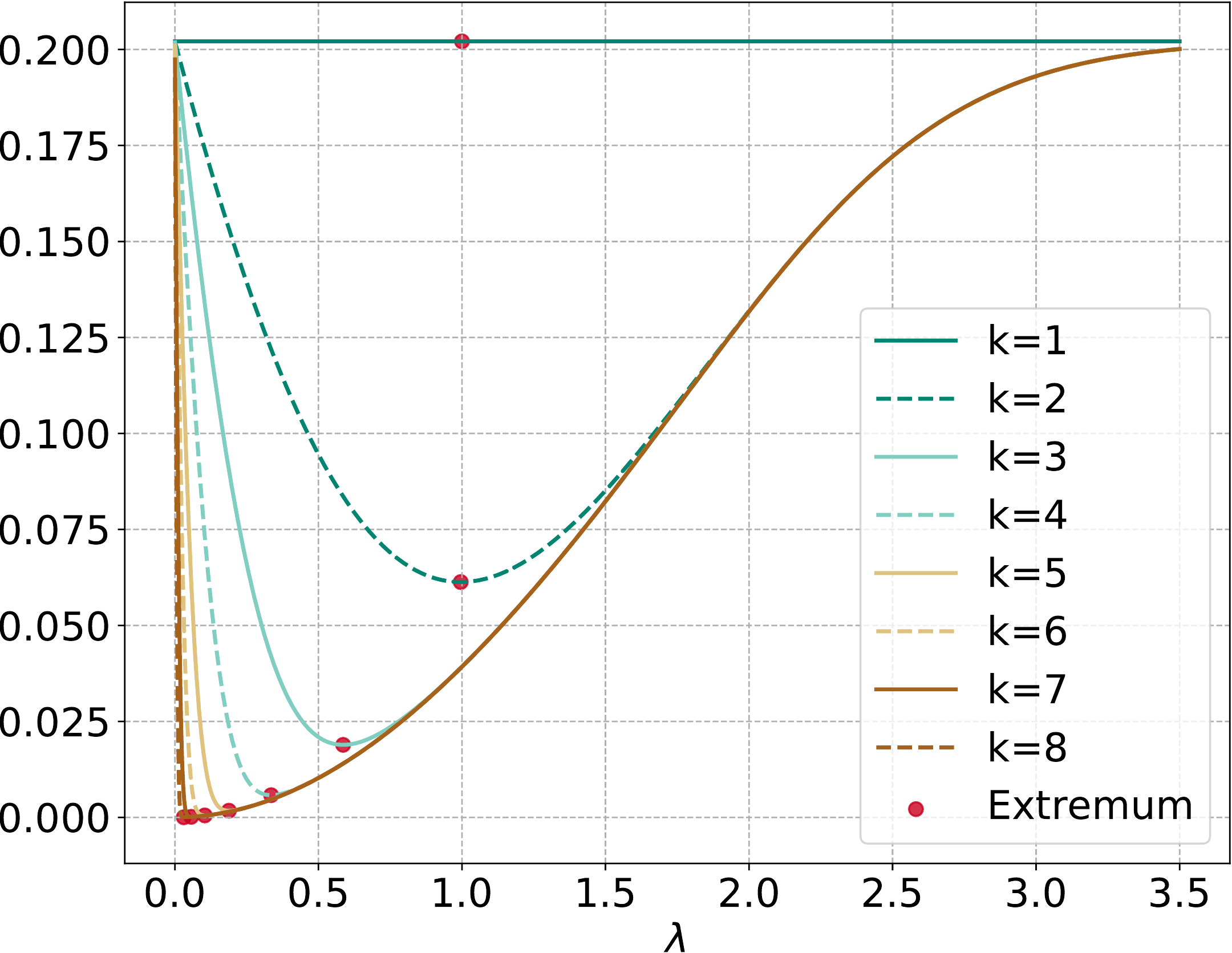}
    \put(-6,35){\rotatebox{90}{\textbf{$J_o(\lambda,k)$}}}
    \end{overpic}
    \vspace{-5pt}
    \caption{The orientation loss
    ($J_{o}(\boldsymbol{\varphi},\boldsymbol{w_q'})\sim~J_o(\lambda,k)$) with different $k$ and $\lambda$ values. The extrema are marked.}
    \label{fig:optimal_orientation_curve}
    \vspace{-6pt}
\end{figure}

\begin{figure*}
    \centering
    \includegraphics[width=0.96\textwidth]{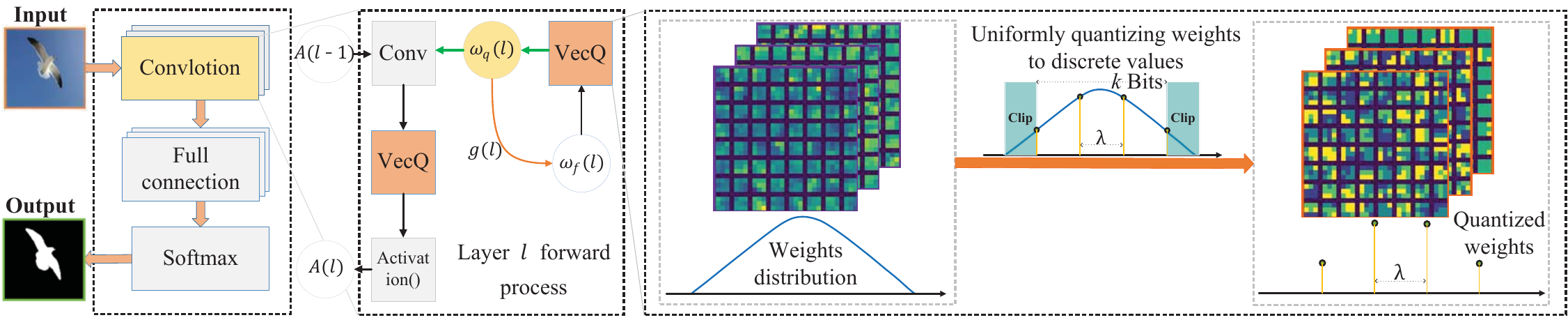}
    \vspace{-5pt}
    \caption{Integrated quantization process in DNN training.}
    \label{fig:voe_quantization_process}
    \vspace{-13pt}
\end{figure*}

Since $\boldsymbol{\varphi} \sim \mathcal{N}(0,1)$, the probability of value $t$ is:
\begin{equation}
    p(t)=\frac{1}{\sqrt{2\pi}} \, \exp \left( -\frac{t^2}{2} \right)
\end{equation}
After the probability $p(t)$ is obtained, the orientation loss function $J_{o}(\boldsymbol{\varphi},\boldsymbol{w_q'})$ can be expressed as a function only relating to $\lambda$ and the targeting bitwidth $k$ for the quantization.
\begin{equation}
\label{eq:jo}
    J_{o}(\boldsymbol{\varphi},\boldsymbol{w_q'}) \sim J_{o}(\lambda,k)
\end{equation}
$J_{o}(\boldsymbol{\varphi},\boldsymbol{w_q'})$ is a convex function with the condition of $k>1$. 
However, it is constant when $k=1$
because the angle between the weight vector and the vector constructed with the signs of values in the weights is constant.
Due to its independence at $k=1$, we set $\lambda$ to $1$ for the convenience of the following process.
We plot the curve of $J_{o}(\lambda,k)$ in \figref{fig:optimal_orientation_curve}~with the change of $\lambda$ under different $k$ bits.

The optimal $\lambda$ values for all bitwidth settings obtained by solving \eqnref{eq:jo}~is shown in \tabref{tab:template_optimal_lambda}. 
The loss is small enough when the targeted bitwidth is greater than 8, so we omit the results for them. 
With the template above,
we only need to solve $J_{o}(\boldsymbol{w_f},\boldsymbol{w_q'})$ once to find the optimal $\lambda$, and then apply it to all quantization without repetitively calculating it.
In other words, simply looking up the corresponding value in this table can obtain the optimal parameter thus reducing the complexity and intensity of the computation,
which significantly speeds up the training process.

\renewcommand{\arraystretch}{1.3}
\begin{table}[ht]
    \caption{Optimal value of $\lambda$ for $J_{o}(\boldsymbol{\varphi},\boldsymbol{w_q'})$ with bitwidth $k$}
    \vspace{-5pt}
    \label{tab:template_optimal_lambda}
    \centering
    \setlength{\tabcolsep}{0.9mm}{
    \begin{tabular}{l|ccccccccc}\Xhline{1.2pt}
         $k$& 1&2&3&4&5&6&7&8& $>$ 8 \\\Xhline{1.2pt}
         $\lambda$ & $(0,\infty)$  &  0.9957  &  0.5860  &  0.3352  &  0.1881   & 0.1041  &  0.0569  &  0.0308&$6/{2^k}$\\\Xhline{1.2pt}
    \end{tabular}
    }
    \vspace{-12pt}
\end{table}

\subsection{DNN training integration}\label{sec:integration}

We integrate our VecQ quantization into the DNN training flow for both the weight data and the activation data,
as shown in \figref{fig:voe_quantization_process}.

\textbf{Weight quantization}:
For layer $l$, during the forward propagation, 
we first quantize the weights with full precision ($\boldsymbol{w_f}(l)$) into the quantized values ($\boldsymbol{w_q}(l)$), then use the quantized weights to compute the output ($\boldsymbol{z}(l)$).
During the backward propagation, 
the gradient is calculated with $\boldsymbol{w_q}(l)$ instead of $\boldsymbol{w_f}(l)$ and propagated.
In the final update process, the gradient $\boldsymbol{g}(l)$ of $\boldsymbol{w_q}(l)$ is used to update $\boldsymbol{w_f}(l)$~\cite{zhou2016dorefa}.

\textbf{Activation quantization}:
Inspired by the Batch Normalization (BN) technique, instead of using pre-defined distribution, we compute the distribution parameter of the activation outputs $p(t)$ and update it with Exponential Moving Average. 
During the inference, the distribution parameter is employed as a linear factor to the activation function~\cite{ioffe2015batch}.
The $A(l)$ is the activation output of layer $l$, and $Activation(\cdot)$ is the non-linear activation function following the convolution or fully-connected layers, such as Sigmoid, Tanh, ReLU.
\section{Evaluations}\label{sec:experiments}
We choose Keras v2.2.4 as the baseline DNN training framework~\cite{chollet2015keras}. 
The layers in Keras are rewritten to support our proposed quantization mechanism as presented in \secref{sec:integration}.
Specifically, all the weight data in the DNNs are quantized to the same bitwidth in the evaluation of VecQ, including the first and last layers.
Our evaluations are conducted on two classic tasks:
(1) image classification and (2) salient object detection (SOD). 
The evaluation results for image classification are compared to state-of-the-art results with the same bitwidth configuration and the SOD results are compared to the state-of-the-art solutions that are conducted with the FP32 data type.

\subsection{Classification}
Image classification is the basis of many computer vision tasks, so the classification accuracy of the quantized model is representative for the effectiveness of our proposed solution.

\subsubsection{Evaluation settings}

\textbf{Datasets and DNN models}.
The MNIST, CIFAR10 \cite{krizhevsky2009learning} and ImageNet \cite{deng2009imagenet} datasets are selected for image classification evaluations;
the IMDB movie reviews~\cite{IMDB-move-review} and THUCNews~\cite{sum2016thuctc} for Chinese text datasets are selected for the sentiment and text classification evaluations.
The detailed information of the datasets are listed in \tabref{tab:datasets_attributes} and \tabref{tab:datasets_sentiment}.

\begin{table}[ht]
\centering
\caption{\label{tab:datasets_attributes}The image classification datasets attributes.}
\vspace{-8pt}
\setlength{\tabcolsep}{3.5mm}{\begin{tabular}{l|ccc}
\Xhline{1pt}
Datasets & \multicolumn{1}{l}{MNIST} & \multicolumn{1}{l}{CIFAR10} & \multicolumn{1}{l}{ImageNet} \\\Xhline{1pt}
Image size  & 28$\times$28$\times$1  & 32$\times$32$\times$3 & 224$\times$224$\times$3  \\
\# of Classes  & 10   & 10   & 1000   \\
\# of Images    & 60000  & 50000 & 1281167  \\
\# of Pixels (log10) & 7.67 & 8.19 & 11.29  \\\Xhline{1pt} 
\end{tabular}}
\vspace{-10pt}
\end{table}

\begin{table}[ht]
\centering
\caption{\label{tab:datasets_sentiment}The sentiment and text classification datasets.}
\vspace{-8pt}
\setlength{\tabcolsep}{5.3mm}{\begin{tabular}{l|cc}
\Xhline{1pt}
Datasets & IMDB & THUCNews \\\Xhline{1pt}
Objectives  & Movie reviews  & Text classification \\
\# of Classes  & 2   & 10 \\
\# of samples  & 50000  & 65000 \\ \Xhline{1pt}
\end{tabular}}
\vspace{-10pt}
\end{table} 

\begin{table}[ht]
    \centering
    \caption{\label{tab:imagenet_architecture}The models for ImageNet.}
    \vspace{-8pt}
    \setlength{\tabcolsep}{3.5mm}{\begin{tabular}{l|ccc}\Xhline{1.2pt}
         Models & AlexNet & ResNet-18 & MobileNetV2  \\\Xhline{1.2pt}
        Convs & 5&21&35 \\ 
        DepConvs&-&-& 17 \\
        BNs & 7& 19&52\\
        FCs & 3& 1&1 \\
        Parameters (M) &50.88&11.7&3.54\\ \Xhline{1.2pt}
    \end{tabular}}\\\raggedright
    \footnotemark{Convs indicate the vanilla convolution layers, 
    DepConvs are the depthwise convolution layers \cite{sandler2018mobilenetv2}. 
    BNs stand for the Batch Normalization layers \cite{ioffe2015batch} and 
    FCs are the full-connection layers.}
    \vspace{-8pt}
\end{table}

\begin{table*}[t!]
    \centering
    \caption{The accuracy and model size with different bitwidth targets.}
    \vspace{-8pt}
    \label{tab:vecq_under_bit_width}
    \setlength{\tabcolsep}{3mm}{\begin{tabular}{c|cc|cc|cc|cc|cc}
    \Xhline{1.5pt}
    \multirow{2}{*}{W/A\footnotemark[1]}& \multicolumn{2}{c|}{LeNet5 } & \multicolumn{2}{c|}{VGG-like~\cite{simonyan2014vgg}} & \multicolumn{2}{c|}{Alexnet~\cite{alexnet}} & \multicolumn{2}{c|}{ResNet-18~\cite{he2016resnet}} & \multicolumn{2}{c}{MobileNetV2~\cite{sandler2018mobilenetv2}} \\\cline{2-11}
    &Size(M)&Acc&Size(M)&Acc&Size(M)&Top1/Top5&Size(M)&Top1/Top5&Size(M)&Top1/Top5\\\Xhline{1.5pt}
    32/32&6.35&99.40&20.44&93.49&194.10&60.01/81.90\footnotemark[2]&44.63&69.60/89.24\footnotemark[3]
    &13.50&71.30/90.10\footnotemark[4]\\\Xhline{0.5pt}
    1/32&0.21&99.34&0.67&90.39& 6.21&55.06/77.78& 1.45&65.58/86.24&0.67&53.78/77.07\\
    2/32&\textbf{0.41}&\textbf{99.53}&1.31&92.94&12.27&59.31/81.01& 2.68&68.23/88.10&1.09&64.67/85.24\\
    3/32&0.60&99.48&1.94&93.02&18.33&60.36/82.40& 4.24&68.79/88.45&1.50&69.13/88.35\\
    4/32&0.80&99.47&2.58&93.27&24.39&61.21/82.94& 5.63&69.80/89.11&1.92&71.89/90.38\\
    5/32&1.00&99.47&3.22&93.37&30.45&61.65/83.19& 7.02&69.98/89.15&2.33&71.47/90.15\\
    6/32&1.20&99.49&3.86&93.51&36.51&62.01/83.32& 8.42&69.81/88.97&2.74&72.23/90.61\\
    7/32&1.40&99.48&\textbf{4.49}&\textbf{93.52}&42.57&62.09/83.44& 9.81&70.17/89.09&\textbf{3.16}&\textbf{72.33/90.62}\\
    8/32&1.60&99.48&5.13&93.50&\textbf{48.63}&\textbf{62.22/83.54}&\textbf{11.20}&\textbf{70.36/89.20}&3.57&72.24/90.66\\\Xhline{0.5pt}
    2/8 & 0.41 & 99.43 & 1.31 &  92.46 & 12.27 &58.04/80.21& 2.68 &67.91/88.30& 1.09 &63.34/84.42\\
    4/8 & 0.80 & 99.53 & 2.58 & 93.37 & 24.39  &61.22/83.24& 5.63 &68.41/88.76&1.92 &71.40/90.41\\
    8/8 & 1.60 & 99.44 & 5.13 & 93.55 &48.63&61.60/83.66& 11.20 &69.86/88.90& 3.57 &72.11/90.68\\
    \Xhline{1.5pt}
    \end{tabular}}\\\raggedright
    \footnotemark[1]{W/A denotes the quantizing bits of weights and activation respectively.}\\
    \footnotemark[2]{Results of AlexNet with Batch Normalization layers are cited from \cite{simon2016cnnmodels}.}\\
    \footnotemark[3]{Results of ResNet18 are cited from \cite{gross2016training}.}\\
    \footnotemark[4]{Results are cited from the document of Keras \cite{chollet2015keras}.}
    \vspace{-10pt}
\end{table*}

For MNIST dataset, Lenet5 with \textbf{32C5-BN-MP2-64C5-BN-MP2-512FC-10Softmax} is used,
where \textbf{C} stands for the Convolutional layer and the number in front denotes the output feature channel number and the number behind is the kernal size; 
\textbf{BN} stands for the Batch Normalization layer;
\textbf{FC} represents the Fully-connected layer and the output channel number is listed in front of it;
\textbf{MP} indicates the max pooling layer followed with the size of the pooling kernel.
The mini-batch size is 200 samples and the initial learning rate is 0.01 and it is divided by 10 at epoch 35 and epoch 50 for a total of 55 training epochs. 

For CIFAR10 dataset, a VGG-like network \cite{simonyan2014vgg} with the architectural configuration as  \textbf{64C3-BN-64C3-BN-MP2-128C3-BN-128C3-BN-MP2-256C3-BN-256C3-BN-MP2-1024FC-10Softmax} is selected.
A simple data augmentation which pads 4 pixels on each side and randomly crops the 32$\times$32 patches from the padded image or its horizontal flip is adopted during the training.
Only the original 32 $\times$ 32 images are evaluated in the test phase.
The network is trained with mini-batch size 128 for a total of 300 epoch. The initial learning rate is 0.01 and decays 10 times at epoch 250 and 290.

For the ImageNet dataset, 
we select 3 famous DNN models, which are AlexNet \cite{alexnet}, ResNet-18 \cite{he2016resnet} and MobileNetV2 \cite{sandler2018mobilenetv2}.
ImageNet dataset contains 1000 categories and the size of the image is relatively bigger~\cite{deng2009imagenet,alexnet,simonyan2014vgg,he2016resnet}.
We use the Batch Normalization (BN) layer instead of the original Local Response Normalization (LRN) layer in AlexNet for a fair comparison with \cite{zhu2016TTQ,ENN2017,TSQ2018}.
The numbers of the different layers and the parameter sizes are listed in~\tabref{tab:imagenet_architecture}.

The architecture of the model for IMDB movie review~\cite{IMDB-move-review} sentiment classification and the THUCNews~\cite{sum2016thuctc} text classification are \textbf{128E-64C5-MP4-70LSTM-1Sigmoid} and \textbf{128E-128LSTM-128FC-10Softmax}, where \textbf{E} denotes Embedding layer and the number in front of it represents its dimension; \textbf{LSTM} is the LSTM layer and the number in front is the number of the hidden units.
In addition, we quantize all layers including Embedding layer, Convolutional layer, Fully-connected layer and LSTM layer for these two models. 
Specifically, for the LSTM layer, we quantize the input features, outputs and the weights, but left the intermediate state and activation of each timestamp untouched since the quantization of them will not help to reduce the size of the model.

\textbf{Evaluation Metrics}
The final classification accuracy results on the corresponding datasets are 
taken as the evaluation metrics in image classification tasks as used in many other works \cite{TWNs,ENN2017,TSQ2018,gysel2016hardware}.
Moreover, the Top1 and Top5 classification accuracy results are presented simultaneously on all the models for ImageNet dataset for comprehensive evaluation as used in \cite{simonyan2014vgg,he2016resnet}.

\subsubsection{Bitwidth flexibility}

\begin{table}
    \centering
    \caption{Evaluation results for LSTM based models.}
    \vspace{-8pt}
    \label{tab:class_imdb}
    \setlength{\tabcolsep}{4.6mm}{
    \begin{tabular}{c|cc|cc}
        \Xhline{1.5pt}
         \multirow{2}{*}{W/A}  & \multicolumn{2}{c|}{IMDB} & \multicolumn{2}{c}{THUCNews}  \\\cline{2-5}
         & Size (M) & Acc. & Size (M) & Acc.\\\Xhline{1.5pt}
         32/32 & 10.07 & 84.98\footnotemark[1] & 3.01 & 94.74\\\Xhline{0.5pt}
     	 2/32&0.63	& 85.54 & 0.19& 93.99\\
         4/32&1.26	& \textbf{86.24} & 0.38 & 94.47 \\
         8/32&2.52   & 85.53  &0.75& 94.53 \\\hline
         2/8	& 0.63	& 85.40 & 0.19&94.00 \\
         4/8	&1.26	& 84.67 &  0.38&94.09 \\
         8/8	&2.52	& 85.72 & 0.75&94.43 \\\Xhline{1.5pt}
    \end{tabular}}\\\flushleft
    \vspace{-5pt}
    \footnotemark[1]{The results of full precision model is from \cite{chollet2015keras}.}
    \vspace{-12pt}
\end{table}

VecQ is designed to support a wide range of targeted bitwidths.
We conduct a series of experiments to verify the impact of bitwidth on model accuracy and model size reduction.
The bitwidth in the following evaluations ranges from $1$ to $8$.

The accuracy results and model sizes for the image classification models are shown in \tabref{tab:vecq_under_bit_width}. 
We first discuss weight quantization only.
There is a relatively higher accuracy degradation when the bitwidth is set to 1.
But starting from 2 bits and up, the accuracy of the models recovers to less than $1.37\%$ drop when compared to the FP32 version except MobileNetV2.
With the increase of bitwidth, the accuracy of the quantized model is improved. 
The highest accuracy of LeNet-5 and VGG-like are $\textbf{99.53\%}$ and $\textbf{93.52\%}$ at 2 bits and 7 bits, respectively, which outperform the accuracy results with FP32. 
The highest accuracy of AlexNet, ResNet-18 and MobileNetV2 are obtained at 8-bit with $\textbf{62.22\%}$ (Top1), 8-bit with $\textbf{70.36\%}$ (Top1) and 7-bit with $\textbf{72.33\%}$ (Top1), respectively, and all of them outperform the values obtained with FP32.
Overall, the best accuracy improvement of the models when compared to the full precision versions for the five models are $\textbf{0.13\%}$, $\textbf{0.03\%}$, $\textbf{2.21\%}$, $\textbf{0.76\%}$ and $\textbf{1.03\%}$, at 2, 7, 8, 8 and 7 bits, respectively when activation is maintained as 32 bits.
The table also contains the accuracy of the models with 8-bit activation quantization. Although the activation quantization leads to a degradation of the accuracy for most of the models (except VGG-like), the results are in general comparable with the models with FP32 data type.

The accuracy and model size of the sentiment classification and text classification models are shown in~\tabref{tab:class_imdb}. Our solution easily outperforms the models trained with FP32. Even with the activation quantization, the accuracy results are still well maintained.
The results also indicate that adopting appropriate quantization of the weight data improves the ability of generalization of the DNN models. 
In another word, right quantization achieves higher classification accuracy for the tests.

\subsubsection{Comparison with State-of-the-art results}
\begin{table}[]
\centering
\caption{\label{tab:comparison_methods_config}
Detailed settings of the quantization methods collected from the literature.}
\vspace{-5pt}
\resizebox{0.49\textwidth}{110pt}{
\setlength\tabcolsep{3pt} 
\begin{tabular}{l|ccc|ccc|ccc}\Xhline{1.5pt}
\multicolumn{1}{c|}{\multirow{2}{*}{\textbf{Methods}}} & \multicolumn{3}{c|}{\textbf{Weights}} & \multicolumn{3}{c|}{\textbf{Activation}} & \multicolumn{1}{c}{\multirow{2}{*}{\textbf{FConv}}} & \multicolumn{1}{c}{\multirow{2}{*}{\textbf{IFC}}} & \multicolumn{1}{c}{\multirow{2}{*}{\textbf{LFC}}} \\\cline{2-7}
\multicolumn{1}{c|}{} & \multicolumn{1}{c}{\textbf{Bits}} & \multicolumn{1}{c}{\textbf{SFB}} & \multicolumn{1}{c|}{\textbf{SFN}} & \multicolumn{1}{c}{\textbf{Bits}} & \multicolumn{1}{c}{\textbf{SFB}} & \multicolumn{1}{c|}{\textbf{SFN}} & \multicolumn{1}{c}{} & \multicolumn{1}{c}{} & \multicolumn{1}{c}{} \\\Xhline{1.5pt}
ReBNet~\cite{ghasemzadeh2018rebnet} & 1 & 32 & 1 & 3 & 32 & 3 & - & Y & N \\
BC~\cite{Binaryconnect} & 1 & - & 0 & 32 & - & - & - & - & - \\
BWN~\cite{rastegari2016xnor} & 1 & 32 & 1 & 32 & - & - & - & - & - \\
BPWN~\cite{TWNs} & 1 & 32 & 1 & 32 & - & - & - & N & N \\
TWN~\cite{TWNs} & 2 & 32 & 1 & 32 & - & - & - & N & N \\
TC~\cite{Ternaryconnect} & 2 & - & 0 & 32 & - & - & - & - & - \\
TNN~\cite{alemdar2017ternary} & 2 & - & 0 & 2 & - & 0 & - & - & - \\
TTQ~\cite{zhu2016TTQ} & 2 & 32 & 2 & 32 & - & - & N & Y & N \\
INQ~\cite{INQ2017} & 2,3,4,5 & - & 0 & 32 & - & - & - & - & - \\
FP~\cite{gysel2016hardware} & 2,4,8 & - & 0 & 32 & - & - & - & N & N \\
uL2Q~\cite{cheng2019uL2Q} & 1,2,4,8 & 32 & 1 & 32 & - & - & Y & Y & 8 \\
ENN~\cite{ENN2017} & 1,2,3 & 32 & 1 & 32 & - & - & - & - & - \\
TSQ~\cite{TSQ2018} & 2 & 32 & c\footnotemark[4] & 2 & 32 & 2 & N & Y & N \\
DC~\cite{han2015deep}\footnotemark[2] & 2,3,4 & 32 & 4,8,16 & 32 & - & - & 8 & - & 1 \\
HAQ~\cite{wang2019haq}\footnotemark[3] & flexible & 32 & 1 & 32 & - & - & 8 & - & 1 \\
QAT~\cite{QAT} & 8 & 32 & 1 & 8 & 32 & 1 & N & - & N \\
TQT~\cite{TQT} & 8 & - & 1 & 8 & - & 1 & 8 & - & 8 \\\Xhline{0.5pt}
\textbf{VecQ} & \textbf{1-8} & \textbf{32} & \textbf{1} & \textbf{32,8} & \textbf{-,32} & \textbf{-,1} & \textbf{Y} & \textbf{Y} & \textbf{Y}\\\Xhline{1.5pt}
\end{tabular}}\\\raggedright
\footnotemark[1]{
\textbf{Weights} and \textbf{Activation} denote the quantized data of the model. 
\textbf{Bits} refer to quantized bitwidth of methods; \textbf{SFB} is the bitwidth for the scaling-factor; \textbf{SFN} is the number of the scaling-factors. \textbf{FConv}, \textbf{IFC} and \textbf{LFC} represent whether the \textbf{F}irst \textbf{Conv}olution layer, the \textbf{I}nternal \textbf{F}ully-\textbf{C}onnected layers and the \textbf{L}ast \textbf{F}ully-\textbf{C}onnected layer are quantized.} \\
\footnotemark[2]{The results of DC are from~\cite{wang2019haq}.} \\
\footnotemark[3]{HAQ is a mix-precision method; results here are from the experiments that only quantize the weight data.} \\
\footnotemark[4]{TSQ introduces a floating-point scaling factor for each convolutional kernel, so the SFN equals to the number of kernels.}
\vspace{-10pt}
\end{table}

We collected the state-of-the-art accuracy results of DNNs quantized to different bitwidth with different quantization methods, and compared them to the results from VecQ with the same bitwidth target. 
The detailed bitwidth support of the comparison methods are listed in \tabref{tab:comparison_methods_config}.
Note here, when the quantization of the activations are not enabled, the SFB and SFN are not applicable to VecQ.

The comparisons are shown in \tabref{tab:imagenet_results}. 
The final accuracy based on VecQ increased by up to 0.62\% , 3.80\%~for LeNet-5 and VGG-like when compared with other quantization methods, respectively.
There is also up to 3.26\%/2.73\%~(top1/top5) improvement of the accuracy of AlexNet, 4.84\%/2.84\%~of ResNet-18 
and 6.60\%/4.00\%~of MobileNetV2 when compared to the \sArt~methods.
For all the 3 datasets and 5 models, the quantized models with VecQ achieve higher accuracy than almost all \sArt~methods with the same bitwidth. 
However, when bitwidth target is 1, the quantized models of AlexNet and Resnet-18 based on VecQ perform worse due to the reason that we have quantized all the weights into 1 bit, including first and last layers, which are different from the counterparts that are using higher bitwidth for the first or last layers. This also leads to more accuracy degradation at low bitwidth on the lightweight network MobiNetV2. This, however, allows us to provide an even smaller model size.
Besides, the solution called BWN~\cite{rastegari2016xnor} and ENN~\cite{ENN2017} for AlexNet has $61$M parameters~\cite{rastegari2016xnor}, while ours is $50.88$M because eliminating the layer paddings in the intermediate layers leads to less weights for fully-connected layers.
When compared to TWN, $\mu$L2Q and TSQ, VecQ achieves significantly higher accuracy at the same targeted bitwidth, which also indicates that our vectorized quantization is superior to the L2 loss based solutions.

\renewcommand{\arraystretch}{1.1}
\begin{table*}[]
    \centering
    \caption{The comparison with other state-of-the-art quantization solutions.}
    \vspace{-8pt}
    \label{tab:imagenet_results}
    \setlength{\tabcolsep}{4.7mm}{
    \begin{tabular}{c|c|c|c|c|c|c}\Xhline{1.5pt}  \multirow{3}{*}{\textbf{Bitwidth}}&\multirow{3}{*}{\textbf{Methods}}&\multicolumn{5}{c}{\textbf{Datasets\&Models}}\\\cline{3-7}
    &&\multicolumn{1}{c|}{\textbf{MNIST}}&\multicolumn{1}{c|}{\textbf{Cifar10}~\cite{krizhevsky2009learning}}&\multicolumn{3}{c}{\textbf{ImageNet}~\cite{deng2009imagenet}}\\\cline{3-7}
    &&\multicolumn{1}{c|}{\textbf{LeNet5}}&\multicolumn{1}{c|}{\textbf{VGG-like}~\cite{simonyan2014vgg}}&\multicolumn{1}{c|}{\textbf{AlexNet}~\cite{alexnet}}&\multicolumn{1}{c|}{\textbf{ResNet18}~\cite{he2016resnet}}&\multicolumn{1}{c}{\textbf{MobileNetV2}~\cite{sandler2018mobilenetv2}}\\\Xhline{1.5pt}
    \textbf{32}&\textbf{FP32}&99.4&93.49&60.01/81.90&69.60/89.24&71.30/90.10\\\Xhline{1.0pt}
    \multirow{8}{*}{1}&{ReBNet\cite{ghasemzadeh2018rebnet}}&98.25&86.98&41.43/-&-&-\\
    &{BC~\cite{Binaryconnect}}&98.82&-&-&-&-\\
    &{BWN\cite{rastegari2016xnor}}&-&-&56.80/79.40&60.80/83.00&-\\
    &{BPWN\cite{TWNs}}&99.05&-&-&57.50/81.20&-\\
    &{ENN~\cite{ENN2017}}&-&-&57.00/79.70&64.80/86.20&-\\
    &{$\mu$L2Q~\cite{cheng2019uL2Q}}&99.06&89.02&-&66.24/86.00&-\\
    &\textbf{VecQ}&\textbf{99.34}&\textbf{90.39}&55.06/77.78&65.58/86.24&53.78/77.07\\\cline{2-7}
    &\textbf{Mean-Imp\footnotemark[3]}& 0.55& 2.39 & 3.32/-1.77&3.24/2.14&-/-
    \\\Xhline{1.0pt}
    \multirow{12}{*}{2}&{FP~\cite{gysel2016hardware}}&98.90&-&-&-&-\\
    &{TWN~\cite{TWNs}}&99.35&-&57.50/79.80\footnotemark[1]&61.80/84.20&-\\
    &{TC~\cite{Ternaryconnect}}&98.85&-&-&-&-\\
    &{TNN~\cite{alemdar2017ternary}}&98.33&-&-&-&-\\
    &{TTQ~\cite{zhu2016TTQ}}&-&-&57.50/79.70&66.60/87.20&-\\
    &{INQ~\cite{INQ2017}}&-&-&-&66.02/87.13&-\\
    &{ENN}&-&-&58.20/80.60&67.00/87.50&-\\
    &{TSQ~\cite{TSQ2018}}&-&-&58.00/80.50&-&-\\
    &{$\mu$L2Q}&99.12&89.50&-&65.60/86.12&-\\
    &{DC\footnotemark[4]}&-&-&-&-&58.07/81.24\\
    &\textbf{VecQ}&\textbf{99.53}&\textbf{92.94}&\textbf{59.31/81.01}&\textbf{68.23/88.10}&\textbf{64.67/85.24}\\\cline{2-7}
    &\textbf{Mean-Imp}& 0.62& 3.44 & 1.51/0.86&2.83/1.67&6.60/4.00
    \\\Xhline{1.0pt}
    \multirow{5}{*}{3}&{INQ}&-&-&-&68.08/88.36&-\\
    &{ENN}\footnotemark[2]&-&-&60.00/82.20&68.00/88.30&-\\
    &{DC}&-&-&-&-&68.00/87.96\\
    &\textbf{VecQ}&99.48&93.02&\textbf{60.36/82.40}&\textbf{68.79/88.45}&\textbf{69.13/88.35}\\\cline{2-7}
    &\textbf{Mean-Imp}& -& - & 0.36/0.20&0.75/0.12&1.13/0.39
    \\\Xhline{1.0pt}
    \multirow{6}{*}{4}&{FP}&99.10&-&-&-&-\\
    &{INQ}&-&-&-&68.89/89.01&-\\
    &{$\mu$L2Q}&99.12&89.80&-&65.92/86.72&-\\
    &{DC}&-&-&-&-&71.24/89.93\\
    &\textbf{VecQ}&\textbf{99.47}&\textbf{93.27}&61.21/82.94&\textbf{69.80}/\textbf{89.11}&\textbf{71.89/90.38}\\\cline{2-7}
    &\textbf{Mean-Imp}& 0.36& 3.47 & -/-&2.39/1.24&0.65/0.45
    \\\Xhline{1.0pt}
    \multirow{3}{*}{5}&{INQ}&-&-&57.39/80.46&68.98/89.10&-\\
    &\textbf{VecQ}&99.47&93.37&\textbf{61.65/83.19}&\textbf{69.98}/\textbf{89.15}&71.47/90.15\\\cline{2-7}
    &\textbf{Mean-Imp}& -& - & 3.26/2.73&1.00/0.05&-/-
    \\\Xhline{1.0pt}
    \multirow{3}{*}{6}
    &{HAQ}&-&-&-&-&66.75/87.32\\
    &\textbf{VecQ}&99.49&93.51&62.01/83.32&69.81/88.87&\textbf{72.23/90.61}\\\cline{2-7}
    &\textbf{Mean-Imp}& -& - & -/-&-/-&5.48/3.29
    \\\Xhline{1.0pt}
    \multirow{1}{*}{7}
    &\textbf{VecQ}&99.48&93.52&62.09/83.44&70.17/89.09&72.33/90.62
    \\\Xhline{1.0pt}
    \multirow{8}{*}{8}&{FP}&99.10&-&-&-&-\\
    &{$\mu$L2Q}&99.16&89.70&-&65.52/86.36&-\\
    &{QAT-c\footnotemark[5]}&-&-&-&-&71.10/-\\
    &{TQT-wt-th}&-&-&-&-&71.80/90.60\\
    &\textbf{VecQ}&\textbf{99.48}&\textbf{93.50}&62.22/83.54&\textbf{70.36/89.20}&\textbf{72.24/90.66}\\\cline{2-7}
    &\textbf{Mean-Imp}& 0.35& 3.80 & -/-&4.84/2.84&0.79/0.06
    \\\Xhline{1.5pt}
    \end{tabular}}\\\raggedright
    \footnotemark[1]{The results of TWN on AlexNet are from \cite{ENN2017}.}\\
    \footnotemark[2]{ENN adopts 2 bits shifting results noted as $\{-4,+4\}$~\cite{ENN2017}.}\\
    \footnotemark[3]{Mean-Imp denotes \textbf{Mean} of accuracy \textbf{Imp}rovement results compared to the state-of-the-art methods.}\\
    \footnotemark[4]{The results are from HAQ~\cite{wang2019haq}.}\\
    \footnotemark[5]{The results are from \cite{QAT_results}.}
    \vspace{-7pt}
\end{table*}

\subsubsection{Analysis of $\lambda$ values}

\begin{table}[!ht]
    \caption{Optimal value of $\lambda$ for $J_{o}(\boldsymbol{\varphi},\boldsymbol{w_q'})$ with bitwidth $k$.}
    \vspace{-8pt}
    \label{tab:theo_prac_lambda}
    \centering
    \setlength{\tabcolsep}{0.8mm}{
    \begin{tabular}{l|ccccccc}\Xhline{1.2pt}
         $k$&2&3&4&5&6&7&8\\\Xhline{1.2pt}
         AlexNet & 0.9700&0.5700&0.3300 &0.1900 &0.1000& 	0.0600&0.0300 \\
         ResNet18& 1.0000 &0.6100 &0.3600 &0.2100 &0.1200&0.0600&	0.0300 \\
         MobileNetV2 & 1.0000 &	0.5900 &	0.3400 &	0.1900 &0.1100 &0.0600 &	0.0300 \\\Xhline{0.5pt}
         MeanError & 0.0171 & -0019   & -0.0024  &  -0.0256 & -0.0178  &  -0.0094 &  0.0023\\\Xhline{1.2pt}
    \end{tabular}
    }
    \vspace{-7pt}
\end{table}

In order to evaluate the accuracy of our theoretical $\lambda$ in~\tabref{tab:template_optimal_lambda}, we choose the last convolutional layers from different models to calculate the actual $\lambda$ values.
Since $\lambda$ is the quantization interval, the range of (0,3] of it covers more than 99.74\% of the layer data, so the actual $\lambda$ is obtained by exhaustively searching the values in the range of (0,3] with the precision at 0.001. 
The comparison is shown in~\tabref{tab:theo_prac_lambda}.
As we can learn from~\tabref{tab:theo_prac_lambda}, there are differences between the theoretical $\lambda$s and the actual values. 
However, the final results in terms of accuracy in the previous subsection is maintained and not effected by the small bias of $\lambda$, which proves the effectiveness of our solution.
\subsection{Salient object detection}
Salient object detection aims at standing out the region of the salient object in an image.
It is an important evaluation that provides good visible results.
Previous experiments show that 2 bits can achieve a good trade-off between accuracy and bitwidth reduction. In this section, only 2 bit quantization for the weights with VecQ is used as the quantization method for the DNN models.

\subsubsection{Evaluation settings}

\textbf{Datasets}
All models are trained with the training data in the MSRA10K dataset (80\% of the entire dataset)~\cite{SOD_survey}.
After training, the evaluation is conducted on multiple datasets, including the MSRA10K (20\% of the entire dataset), ECSSD \cite{SOD_survey}, HKU-IS \cite{SOD_survey},
DUTs \cite{SOD_survey}, DUT-OMRON \cite{SOD_survey}
and the images containing target objects and existing ground truth maps in the THUR15K \cite{ChengGroupSaliency}. 
The details of the selected datasets are shown in \tabref{tab:sod_datasets}. 
All images are resized to $224 \times 224$ for the training and test.

\begin{table}[ht]
\centering
\caption{\label{tab:sod_datasets}The datasets for salient object detection.}
\vspace{-8pt}
\setlength{\tabcolsep}{6.2mm}{\begin{tabular}{l|cc}
\Xhline{1.2pt}
\textbf{Datasets} & \textbf{Images} & \textbf{Contrast} \\\Xhline{1.2pt}
\textbf{ECSSD }&  1000 & High \\
\textbf{HKU-IS } & 4000 & Low \\
\textbf{DUTs } & 15572 & Low\\
\textbf{DUT-OMRON } & 5168& Low\\
\textbf{MSRA10K } & 10000 (80/20\%) & High\\
\textbf{THUR15K } & 6233 & Low\\
\Xhline{1.2pt} 
\end{tabular}}
\vspace{-14pt}
\end{table}

\textbf{Models}
~The famous end-to-end semantic segmentation models i.e., U-Net \cite{SOD_survey},
FPN \cite{lin2017fpn}, LinkNet \cite{chaurasia2017linknet} and UNet++ \cite{unetplusplus} are selected for the comprehensive comparison. Their detailed information is shown in \tabref{tab:sod_models}. 
The models are based on the same ResNet-50 backbone as encoder and initialized with the weights trained on the ImageNet dataset.

\begin{table}[ht]
\centering
\caption{\label{tab:sod_models}The models for salient object detection.}
\vspace{-8pt}
\setlength{\tabcolsep}{1.8mm}{
\begin{tabular}{l|ccccc}
\Xhline{1.2pt}
\textbf{Models} & \textbf{U-Net} & \textbf{FPN} & \textbf{LinkNet} & \textbf{UNet++} \\\Xhline{1.2pt}
\textbf{Backbone}& ResNet-50 & ResNet-50 &ResNet-50 & ResNet-50 \\
\textbf{Convs} &64&67&69&76\\
\textbf{Parameters (M)}&36.54&	28.67&	28.78&	37.7\\
\textbf{Model size (M)}&139.37	&109.38	&109.8&	143.81\\
\textbf{Q. size (M)\footnotemark[1]}&\textbf{9.05}	&\textbf{7.20}	&\textbf{7.24}&	\textbf{9.35}
\\\Xhline{1.2pt}                
\end{tabular}
}\\\flushleft
\vspace{-5pt}
\footnotemark[1]{Q. size (M) stands for the size of the VecQ quantized model.}
\vspace{-5pt}
\end{table}


\begin{table*}[t!]
\centering
\caption{The salient object detection results.}
\vspace{-6pt}
\label{tab:sod_results}
\resizebox{1.0\textwidth}{27mm}{
\setlength\tabcolsep{1pt} 
\begin{tabular}{l|l||cccc|cccc|cccc|cccc|cccc|cccc}\Xhline{1.2pt}
                  & \textbf{Datasets} & \multicolumn{4}{c|}{\textbf{MSRA10K-test}}                            & \multicolumn{4}{c|}{\textbf{ECSSD}}                                   & \multicolumn{4}{c|}{\textbf{HKU-IS}}                                  & \multicolumn{4}{c|}{\textbf{DUTs}}                                    & \multicolumn{4}{c|}{\textbf{DUT-OMRON}}                               & \multicolumn{4}{c}{\textbf{THUR15K}}                                   \\\Xhline{1.2pt}
\textbf{Model}    & \textbf{Size (M)} & \textbf{MAE $\downarrow$}    & \textbf{MaxF $\uparrow$}   & \textbf{S $\uparrow$}      & \textbf{E $\uparrow$}      & \textbf{MAE $\downarrow$}    & \textbf{MaxF $\uparrow$}   & \textbf{S $\uparrow$}      & \textbf{E $\uparrow$}     & \textbf{MAE $\downarrow$}    & \textbf{MaxF $\uparrow$}   & \textbf{S $\uparrow$}      & \textbf{E $\uparrow$}     & \textbf{MAE $\downarrow$}    & \textbf{MaxF $\uparrow$}   & \textbf{S $\uparrow$}      & \textbf{E $\uparrow$}     & \textbf{MAE $\downarrow$}    & \textbf{MaxF $\uparrow$}           & \textbf{S $\uparrow$}              & \textbf{E  $\uparrow$}              & \textbf{MAE $\downarrow$}             &  \textbf{MaxF $\uparrow$}            & \textbf{S $\uparrow$}             & \textbf{E  $\uparrow$  }            \\\Xhline{1.2pt}
\textbf{Unet}                        & 139.37                      & 0.030                                 & 0.945                                  & 0.931                                  & 0.962                                  & 0.057                                 & 0.909                                  & 0.886                                  & 0.914          & 0.045                                 & 0.907                                  & 0.884                                  & 0.930          & 0.060                                 & 0.896                                  & 0.865                                  & 0.874          & 0.070                                 & 0.804          & 0.803                                  & 0.829                                  & 0.077                                  & 0.769         & 0.807                                  & 0.816                                  \\
\textbf{Unet*}                       & \emph{ 9.05} & 0.032                                 & 0.940                                  & 0.926                                  & 0.959                                  & 0.064                                 & 0.896                                  & 0.871                                  & 0.906          & 0.050                                 & 0.893                                  & 0.870                                  & 0.923          & 0.065                                 & 0.885                                  & 0.852                                  & 0.871          & 0.071                                 & 0.793          & 0.795                                  & 0.835                                  & 0.081                                  & 0.749         & 0.797                                  & 0.815                                  \\\Xhline{0.5pt}
\textbf{Bias}                        & 93.51\%                      & \textbf{-0.002}                       & \textbf{0.004}                         & \textbf{0.005}                         & \textbf{0.003}                         & \textbf{-0.007}                       & 0.013                                  & 0.015                                  & \textbf{0.008} & \textbf{-0.005}                       & 0.014                                  & 0.015                                  & \textbf{0.007} & \textbf{-0.005}                       & 0.011                                  & 0.012                                  & \textbf{0.003} & \textbf{-0.001}                       & 0.011          & \textbf{0.008}                         & \emph{\textbf{-0.006}} & \textbf{-0.005}                        & 0.020         & \textbf{0.010}                         & \textbf{0.001}                         \\\Xhline{1.2pt}
\textbf{FPN}                         & 109.38                      & 0.043                                 & 0.920                                  & 0.899                                  & 0.949                                  & 0.070                                 & 0.882                                  & 0.854                                  & 0.901          & 0.059                                 & 0.875                                  & 0.848                                  & 0.911          & 0.072                                 & 0.875                                  & 0.835                                  & 0.861          & 0.081                                 & 0.777          & 0.772                                  & 0.812                                  & 0.087                                  & 0.750         & 0.778                                  & 0.804                                  \\
\textbf{FPN*}                        & \emph{ 7.2}  & 0.038                                 & 0.935                                  & 0.920                                  & 0.955                                  & 0.070                                 & 0.889                                  & 0.866                                  & 0.897          & 0.059                                 & 0.879                                  & 0.859                                  & 0.908          & 0.070                                 & 0.878                                  & 0.850                                  & 0.859          & 0.080                                 & 0.772          & 0.786                                  & 0.809                                  & 0.089                                  & 0.739         & 0.789                                  & 0.796                                  \\\Xhline{0.5pt}
\textbf{Bias} & 93.42\%                      & \emph{\textbf{0.005}} & \emph{\textbf{-0.015}} & \emph{\textbf{-0.022}} & \emph{\textbf{-0.006}} & \emph{\textbf{0.000}} & \emph{\textbf{-0.007}} & \emph{\textbf{-0.012}} & \textbf{0.004} & \emph{\textbf{0.001}} & \emph{\textbf{-0.005}} & \emph{\textbf{-0.011}} & \textbf{0.004} & \emph{\textbf{0.002}} & \emph{\textbf{-0.003}} & \emph{\textbf{-0.015}} & \textbf{0.003} & \emph{\textbf{0.001}} & \textbf{0.004} & \emph{\textbf{-0.014}} & \textbf{0.003}                         & \emph{\textbf{-0.002}} & 0.011         & \emph{\textbf{-0.011}} & \textbf{0.007}    
 \\\Xhline{1.2pt}
\textbf{Linknet}                     & 109.8                       & 0.032                                 & 0.942                                  & 0.928                                  & 0.959                                  & 0.060                                 & 0.905                                  & 0.882                                  & 0.911          & 0.048                                 & 0.900                                  & 0.878                                  & 0.927          & 0.062                                 & 0.892                                  & 0.861                                  & 0.871          & 0.071                                 & 0.801          & 0.799                                  & 0.825                                  & 0.079                                  & 0.760         & 0.803                                  & 0.814                                  \\
\textbf{Linknet*}                    & \emph{ 7.24} & 0.034                                 & 0.939                                  & 0.923                                  & 0.959                                  & 0.068                                 & 0.891                                  & 0.865                                  & 0.902          & 0.054                                 & 0.887                                  & 0.860                                  & 0.920          & 0.068                                 & 0.883                                  & 0.847                                  & 0.870          & 0.072                                 & 0.788          & 0.787                                  & 0.833                                  & 0.082                                  & 0.746         & 0.794                                  & 0.818                                  \\\Xhline{0.5pt}
\textbf{Bias}                        & 93.40\%                      & \textbf{-0.002}                       & \textbf{0.003}                         & \textbf{0.004}                         & \textbf{0.001}                         & \textbf{-0.008}                       & 0.014                                  & 0.017                                  & \textbf{0.008} & \textbf{-0.005}                       & 0.013                                  & 0.018                                  & \textbf{0.006} & \textbf{-0.005}                       & 0.010                                  & 0.014                                  & \textbf{0.001} & \textbf{-0.001}                       & 0.014          & 0.012                                  & \emph{\textbf{-0.008}} & \textbf{-0.002}                        & 0.014         & \textbf{0.009}                         & \emph{\textbf{-0.004}} \\\Xhline{1.2pt}
\textbf{UNet++}                      & 143.81                      & 0.029                                 & 0.948                                  & 0.933                                  & 0.964                                  & 0.056                                 & 0.910                                  & 0.888                                  & 0.915          & 0.044                                 & 0.909                                  & 0.887                                  & 0.930          & 0.059                                 & 0.897                                  & 0.867                                  & 0.876          & 0.070                                 & 0.805          & 0.805                                  & 0.829                                  & 0.076                                  & 0.769         & 0.810                                  & 0.818                                  \\
\textbf{UNet++*}                     & \emph{ 9.35} & 0.033                                 & 0.939                                  & 0.926                                  & 0.958                                  & 0.065                                 & 0.895                                  & 0.872                                  & 0.905          & 0.052                                 & 0.890                                  & 0.868                                  & 0.919          & 0.066                                 & 0.884                                  & 0.854                                  & 0.867          & 0.075                                 & 0.785          & 0.792                                  & 0.822                                  & 0.082                                  & 0.750         & 0.797                                  & 0.811                                  \\\Xhline{0.5pt}
\textbf{Bias}                        & 93.50\%                      & \textbf{-0.004}                       & \textbf{0.008}                         & \textbf{0.007}                         & \textbf{0.007}                         & \textbf{-0.009}                       & 0.015                                  & 0.016                                  & 0.010          & \textbf{-0.008}                       & 0.019                                  & 0.018                                  & 0.011          & \textbf{-0.007}                       & 0.013                                  & 0.013                                  & \textbf{0.009} & \textbf{-0.005}                       & 0.020          & 0.013                                  & \textbf{0.007}                         & \textbf{-0.006}                        & 0.019         & 0.013                                  & \textbf{0.007}                         \\\Xhline{1.2pt}
\end{tabular}}\\\flushleft
\vspace{-5pt}
\footnotemark[1]{S and E stand for S-measure and E-measure, respectively. The $\uparrow$ indicates that the higher value shows better results and the $\downarrow$ is vise versa. The $*$ indicates the quantized models with VecQ-2 and the quantized model sizes are marked in \emph{italics}. \textbf{Bias} row lists the difference between the full-precision model and the quantized model. A negative value is better in S$\uparrow$, E$\uparrow$, and MaxF$\uparrow$ column but worse in the MAE$\downarrow$ column}.
\vspace{-6pt}
\end{table*}

\textbf{Evaluation Metrics}
We choose $4$ widely used metrics for a comprehensive evaluation: 
(1) Mean Absolute Error (MAE) \cite{SOD_survey},
(2) Maximal F-measure (MaxF) \cite{SOD_survey},
(3) Structural measure (S-measure) \cite{SOD_survey},
and (4) Enhanced-alignment measure (E-measure) \cite{fan2018enhanced}. 

The MAE measures the average pixel-wise absolute error between the output map $F$ and the ground-truth mask $G$.
\begin{equation}
    MAE=\frac{1}{m}\sum_{k=1}^{m}(\frac{1}{H_k\times W_k}\sum_{i=1}^{H_k}\sum_{j=1}^{W_k}|G_k(i,j)-F_k(i,j)|)
\end{equation}
Here $m$ is the number of the samples, $H_k$ and $W_k$ are the height and width of $G_k$. 

F-measure comprehensively evaluates the $Precision$ and $Recall$ with a weight parameter $\beta$.
\begin{equation}
    F_{\beta}= \frac{(1+\beta^2)Precision+Recall}{{\beta^2}Precision+Recall}
\end{equation}
$\beta^2$ is empirically set to $0.3$ \cite{SOD_survey}. 
The MaxF is the maximal F$_\beta$ value.

S-measure considers the object-aware and region-aware structure similarities.

E-measure is designed for binary map evaluations. It combines the local pixel-wise difference and the global mean value of map for comprehensive evaluation.
In our evaluation, the output map is first binarized to [0,1] by comparing with the threshold of twice of its mean value \cite{SOD_survey}, then the binary map is evaluated with E-measure.

We also involve direct visual comparison of the full precision and quantized weights of the selected model, to provide a more visible comparison.

\vspace{-2pt}
\subsubsection{Results and analysis}
\begin{figure*}[t!]
    \centering
    \includegraphics[width=1\textwidth]{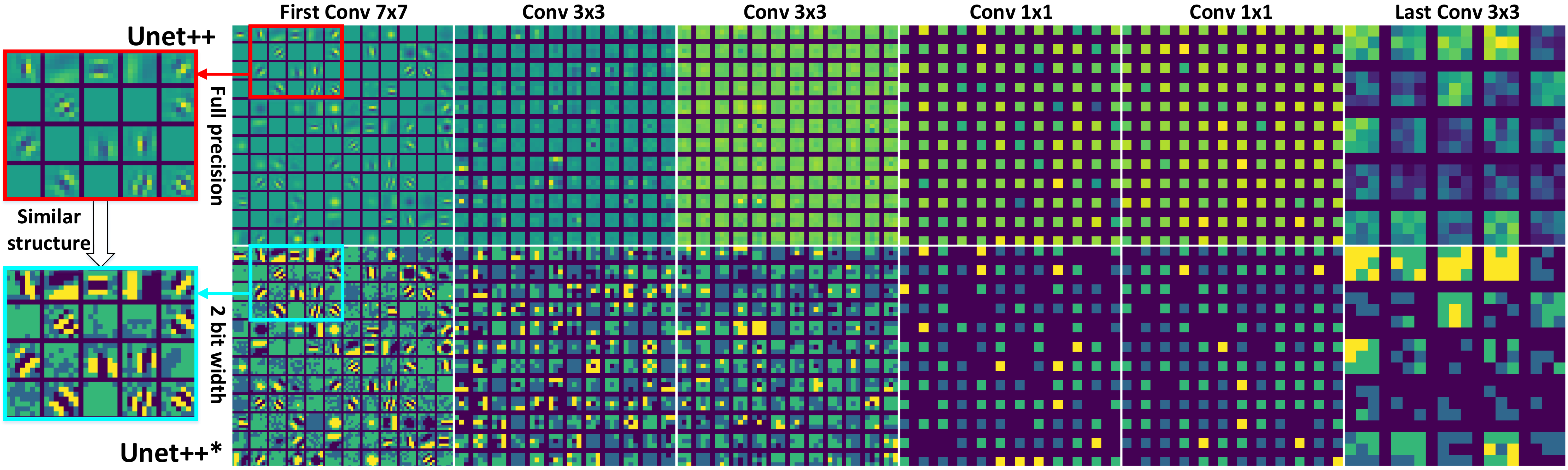}
    \vspace{-20pt}
    \caption{The comparison of weights of convolutional layers in Unet++ and Unet++*.}
    \label{fig:sod_weights_imgs}
    \vspace{-6pt}
\end{figure*}

The 
quantitative comparison results are in \tabref{tab:sod_results}.
The $*$ indicates the quantized model based on VecQ besides the full precision model. 
Note here, the output sizes of the last layer of FPN and FPN* are $56 \times 56$ and then resized to $224 \times 224$.
Overall, the performance degradation of the quantized models based on VecQ is less than 0.01 in most metric for all models, but the size of the quantized model is reduced by more than 93\% when compared to the models using FP32.

As shown in \tabref{tab:sod_results}, all the quantized models have a less than 0.01 degradation on MSRA10K with all evaluation metrics.
This is because all the models are trained on the training subset of MSRA10K, so the features of the images in it are well extracted.
The other 5 data sets are only used as testing datasets and there are more degradation with the evaluation metrics, especially in MaxF and S-measure, but the degradation is maintained within 0.02.

Comparing the quantized models with their corresponding full-precision models, FPN* performs well on almost all the test sets with all the evaluation metrics (shown as \emph{\textbf{bold italics}} numbers in \tabref{tab:sod_results}), showing a better feature extraction capacity than the full precision version (FPN).
Compared to other models, FPN outputs a relatively smaller prediction map ($56\times56$). 
Although the backbone models are similar, the feature classification tasks in the FPN work on a smaller feature map with a similar size of coefficients. 
This provides a good chance for data quantization without effecting the detection ability because of the potential redundancy of the coefficients.

In order to further present the effectiveness of our proposed quantization solution,
we print the weights of the convolution layers of the backbone model in Unet++ and Unet++*, as shown in \figref{fig:sod_weights_imgs}. 
There are three sizes of kernels involved in this model, which are $7\times7$, $3\times3$ and $1\times1$. 
In addition, we also compare the full precision and quantized weights of the last convolution layer for the selected model, which directly outputs the results for the detection.

In the first $7\times7$ kernels, we notice a significant structural similarity between the full precision weights and the quantized weights. 
Since the very first layer of the backbone model extracts the overall features, the quantized weights provide a nice feature extraction ability.
When the size of the kernels become smaller, we could still notice a good similarity between the full precision ones and quantized ones, Although the similarities are not significant in the Conv 3x3 sets, they become obvious in the following Conv 1x1 sets.
The Last Conv group directly explains the comparable output results with the visible emphasized kernels and locations of the weights.

Overall, the quantized weights show a good similarity to the full precision ones in terms of the value and the location which ensures the high accuracy output when compared to the original full precision model.
\vspace{-6pt}
\section{Conclusion}\label{sec:conclusion}
\vspace{-2pt}
In this paper, we propose a new quantization solution called VecQ. 
Different from the existing works, it utilizes the vector loss instead of adopting L2 loss to measure the loss of quantization.
VecQ quantizes the full-precision weight vector into a specific bitwidth with the least DQL and, hence, provides a better final model accuracy.
We further introduce the fast quantization algorithm based on a reasonable prior knowledge of normally distributed weights and reduces the complexity of the quantization process in model training.
The integration of VecQ into Keras \cite{chollet2015keras} is also presented and used for our evaluations.
The comprehensive evaluations have shown the effectiveness of VecQ on multiple datasets, models and tasks. 
The quantized low-bit models based on VecQ show comparable classification accuracy to models with FP32 datatype and outperform all the state-of-the-art quantization methods when the targeted bitwidth of the weights is higher than 2.
Moreover, the experiments on salient object detection also show that VecQ can greatly reduce the size of the models while maintaining the performance of feature extraction tasks.

For future work, we will focus on the combination of non-linear quantization and illustrate an automated mixed-precision quantization with VecQ to achieve better performance improvement.
The source code of the Keras built with VecQ could be found at https://github.com/GongCheng1919/VecQ.
\vspace{-6pt}
\section{Acknowledgment}
\vspace{-4pt}
This work is partially supported by the National Natural Science Foundation (61872200), the National Key Research and Development Program of China (2018YFB2100304, 2018YFB1003405), the Natural Science Foundation of Tianjin (19JCZDJC31600, 19JCQNJC00600), the Open Project Fund of State Key Laboratory of Computer Architecture, Institute of Computing Technology, Chinese Academy of Sciences (CARCH201905).
It is also partially supported by the National Research Foundation, Prime Minister's Office, Singapore under its Campus for Research Excellence and Technological Enterprise (CREATE) programme, and the IBM-Illinois Center for Cognitive Computing System Research (C3SR) - a research collaboration as part of IBM AI Horizons Network.
\vspace{-6pt}

\bibliographystyle{IEEEtran}
\vspace{-5pt}
\bibliography{main}

\vspace{-12mm}
\begin{IEEEbiography}[{\includegraphics[width=1in,height=1.25in,clip]{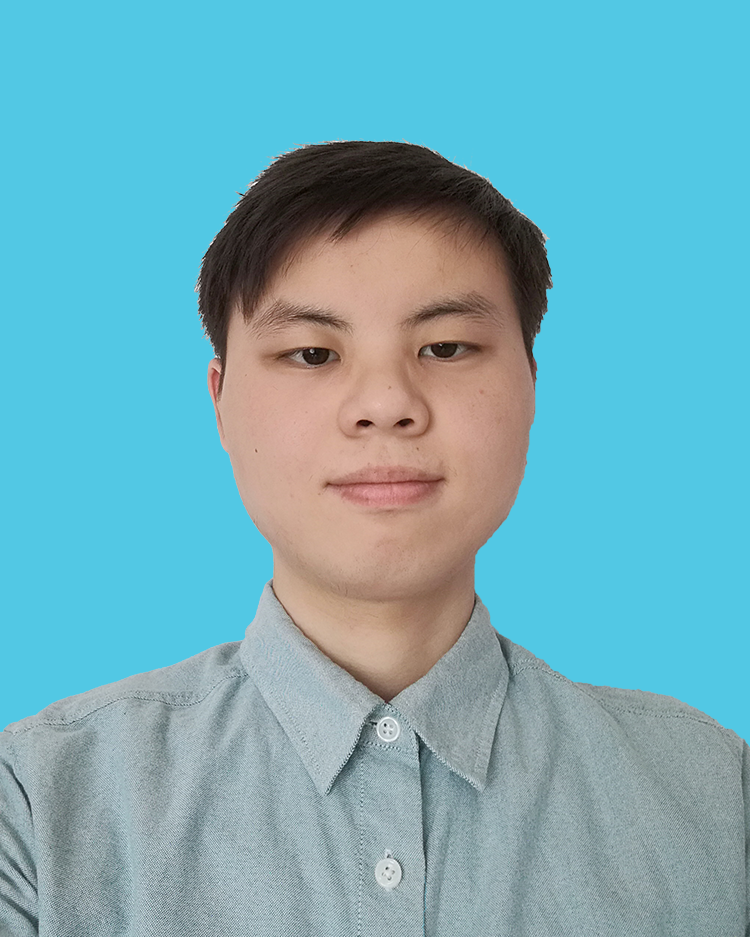}}]{Cheng Gong}
received his B.Eng. degree in computer science from Nankai University in 2016. He is currently working toward his Ph.D. degree in the College of Computer Science, Nankai University. His main research interests include heterogeneous computing, machine learning and Internet of Things.
\end{IEEEbiography}
\vspace{-10mm}
\begin{IEEEbiography}[{\includegraphics[width=1in,height=1.25in,clip]{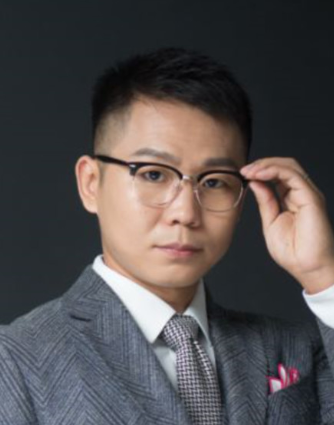}}]{Yao Chen} received the B.S. and Ph.D. degree from Nankai University, Tianjin, China in 2010 and 2016, respectively. He is currently a research scientist in the Advanced Digital Sciences Center, Singapore, which is a research institute of University of Illinois at Urbana-Champaign. His research interests include reconfigurable computing, high level synthesis and high performance computing.
\end{IEEEbiography}
\vspace{-10mm}
\begin{IEEEbiography}[{\includegraphics[width=1in,height=1.25in,clip]{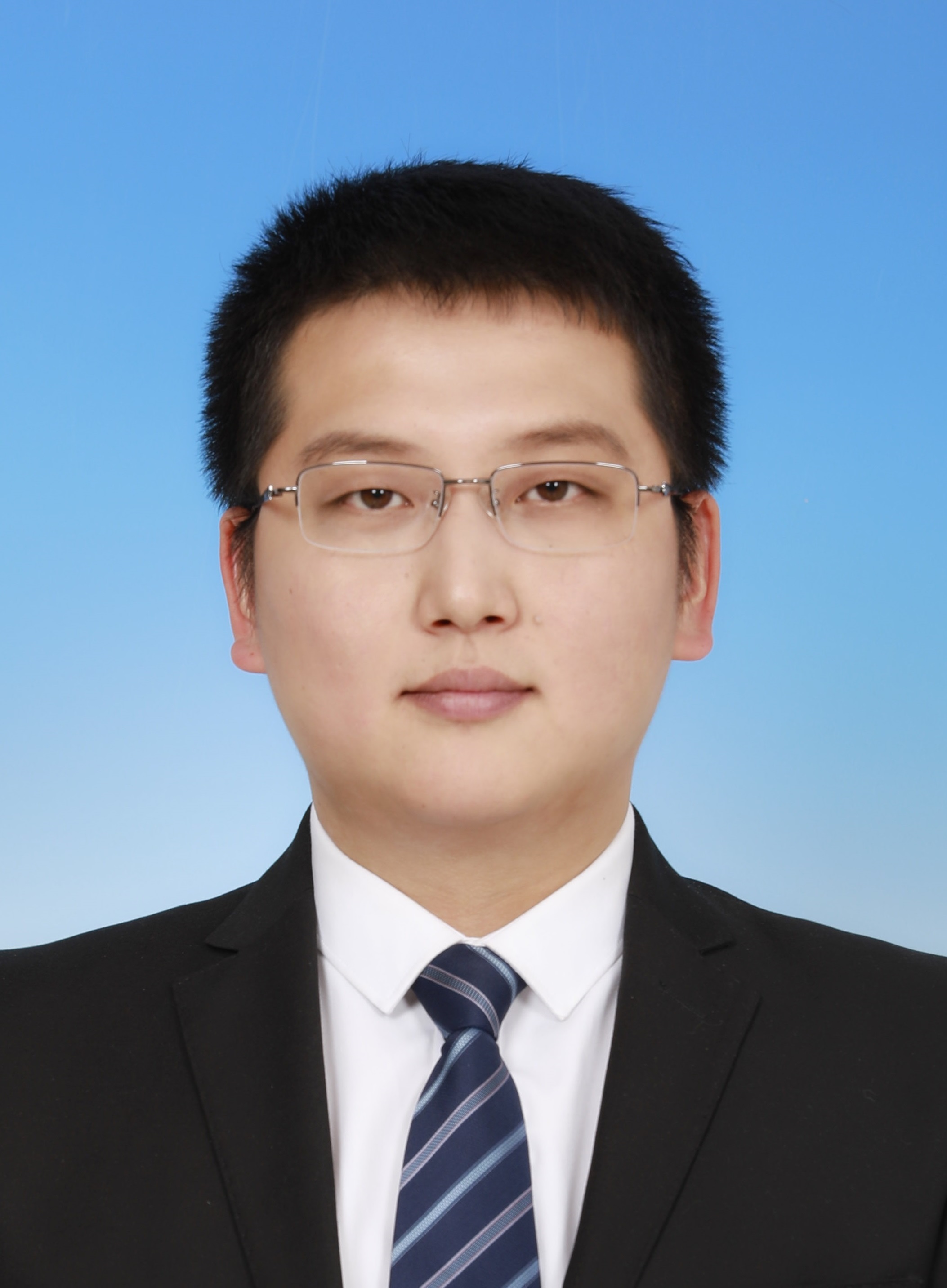}}]{Ye Lu}
received the B.S. and Ph.D. degree from Nankai University, Tianjin, China in 2010 and 2015, respectively. He is an associate professor at the College of Cyber Science, Nankai University now. His main research interests include embedded system, Internet of Things and artificial intelligence.
\end{IEEEbiography}
\vspace{-10mm}
\begin{IEEEbiography}[{\includegraphics[width=1in,height=1.25in,clip]{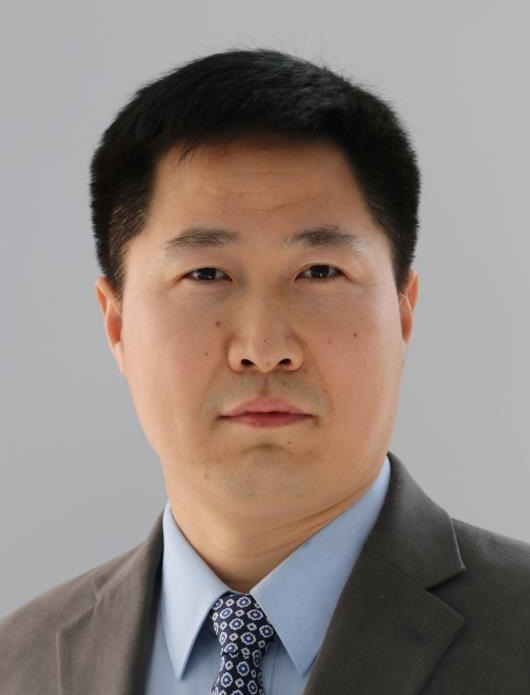}}]{Tao Li}
received his Ph.D. degree in Computer Science from Nankai University, China in 2007. He works at the College of Computer Science, Nankai University as a Professor. He is the Member of the IEEE Computer Society and the ACM, and the distinguished member of the CCF. His main research interests include heterogeneous computing, machine learning and Internet of things.
\end{IEEEbiography}
\vspace{-10mm}
\begin{IEEEbiography}[{\includegraphics[width=1in,height=1.25in,clip]{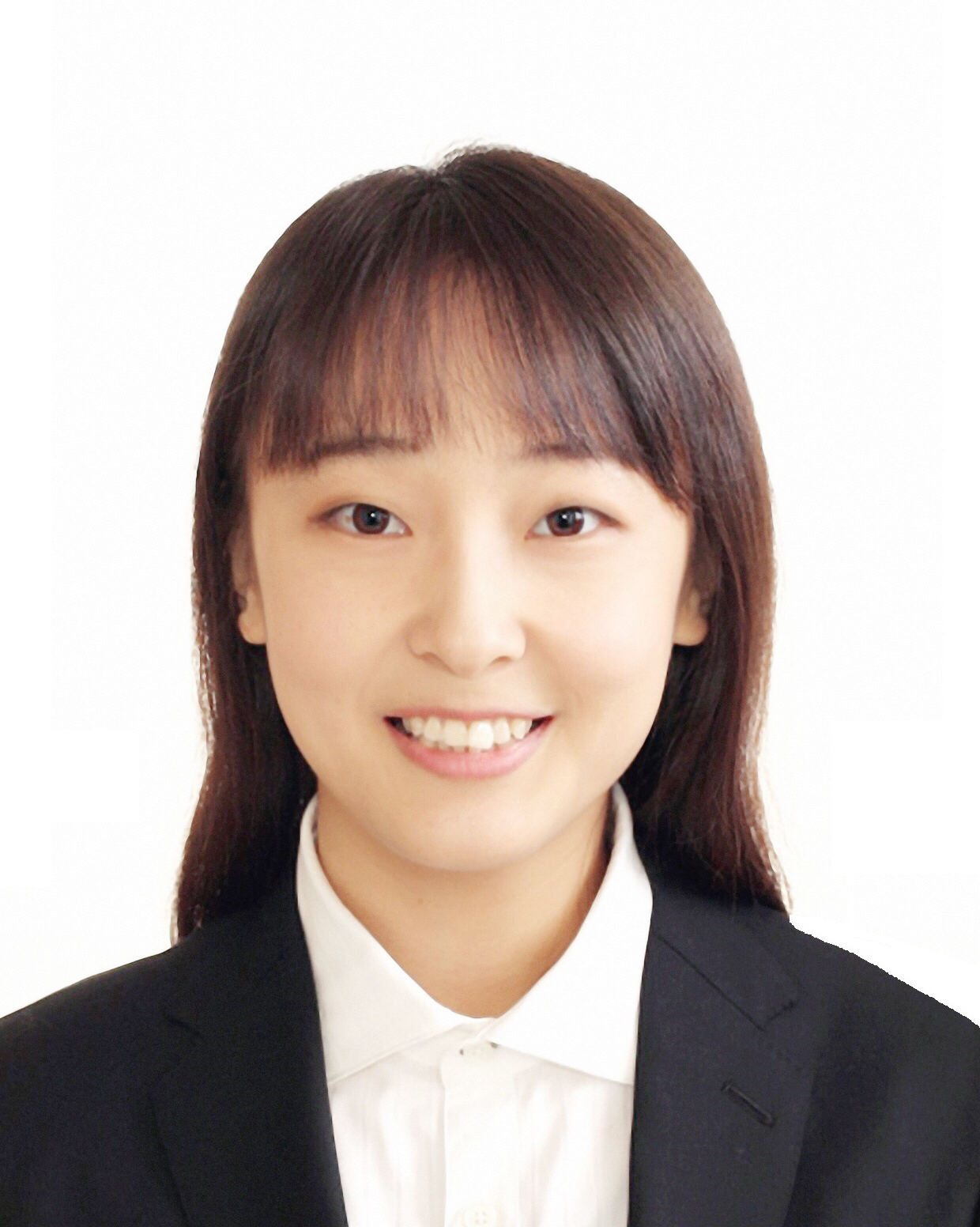}}]{Cong Hao}
received her Ph.D. degree in electronic engineering from Waseda University, Japan, in 2017, and the M.S. and B.S. degrees from Shanghai Jiao Tong University in 2014 and 2011. She is currently a postdoctoral researcher with the ECE Department, University of Illinois at Urbana-Champaign. Her current research interests
include system-level and high-level synthesis, EDA techniques and reconfigurable computing.
\end{IEEEbiography}
\vspace{-10mm}
\begin{IEEEbiography}[{\includegraphics[width=1in,height=1.25in,clip]{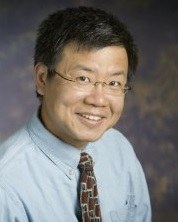}}]{Deming Chen}
received the B.S. degree in computer science from the University of
Pittsburgh, PA, USA, in 1995, and the M.S. and Ph.D. degrees in computer science
from the University of California at Los Angeles, in 2001 and 2005, respectively. He
is the Abel Bliss Endowed Professor 
with the ECE Department, University of Illinois at Urbana–Champaign. Dr. Chen is an IEEE Fellow, an ACM Distinguished Speaker, and the Editor-in-Chief of ACM Transactions on Reconfigurable Technology and Systems (TRETS).
His current research interests include system-level and high-level synthesis, machine learning, GPU and reconfigurable computing, computational genomics, and hardware security.
\end{IEEEbiography}

\end{document}